\documentclass[runningheads]{llncs}

\usepackage{multirow}
\usepackage{amsmath}
\usepackage{makecell}
\usepackage[table]{xcolor}
\usepackage{xcolor}
\usepackage{siunitx}
\usepackage{algorithm}
\usepackage{pifont}        % gives \ding
\usepackage{enumitem}
\usepackage[normalem]{ulem}
\usepackage{wrapfig}

\definecolor{Bred}{HTML}{b5341e}   
\definecolor{Fgreen}{HTML}{309c59}
\newcommand{\cmark}{\textcolor{Fgreen}{\ding{52}}} % ✓  (\ding{51} also works)
\newcommand{\xmark}{\textcolor{Bred}{\ding{56}}}

\usepackage{eccv}

\usepackage{eccvabbrv}
\usepackage{multirow}
\usepackage{graphicx}
\usepackage{booktabs}

\usepackage{array}
\usepackage{adjustbox}
\usepackage{pifont}
\usepackage{gensymb}
\usepackage{bbding}

\usepackage{soul}
\setulcolor{black!30}
\setul{0.3ex}{0.15ex}

\usepackage[accsupp]{axessibility}  % Improves PDF readability for those with disabilities.

\usepackage[pagebackref,breaklinks,colorlinks,citecolor=eccvblue]{hyperref}
\usepackage{orcidlink}

\newcommand{\YX}[1]{{\textcolor{green}{[\textbf{YX:} #1]}}}
\newcommand{\MK}[1]{{\textcolor{blue!70}{[\textbf{MK:} #1]}}}

\newcommand{\smplx}{SMPL({$-$}X)}
\newcommand{\smpld}{$+$D}
\newcommand{\avaimg}{\texttt{AvaImg}}
\newcommand{\todo}[1]{{\color{orange}#1}}

\newcolumntype{P}[1]{>{\centering\arraybackslash}p{#1}}

\newcommand{\edit}[2]{\textcolor{black}{#2}}

\newif\ifincludesupplementary
\includesupplementaryfalse
\begin{document}

% \title{Avatar As Image: Representing High-Fidelity \\ 3D Clothed Human in Images} 
% \title{Avatar As Image: Revisiting 3D Human as Images with High-Fidelity Textured Registration} 
\title{Revisiting Avatar-As-Image: High-Fidelity Registration is All You Need} 
% Avatar to Image: [Achieving Cohesion for] Representing High-Fidelity 3D Clothed Humans as 2D Images
%                                           Representing High-Fidelity 3D Clothed Humans in UV Space ?
% Av2Img instead then 



% [OLD — replaced by Claude 2026-06-20 (shared * for equal contribution, separate dagger for corresponding author)]
% \author{Margaret Kostyrko\textsuperscript{1,2}\orcidlink{0009-0004-2695-4897} \and
% Yuxuan Xue\textsuperscript{1,2}\orcidlink{0000-0001-8521-1293}\thanks{Corresponding Author}\thanks{Equal Contribution} \and
% Garvita Tiwari\textsuperscript{1,2}\orcidlink{0000-0002-3484-942X} \\
% Gerard Pons-Moll\textsuperscript{1,2,4}\orcidlink{0000-0001-5115-7794}}
%
%
% \author{Margaret Kostyrko\inst{1,2}\orcidlink{0009-0004-2695-4897} \and
% Yuxuan Xue\inst{1,2}\orcidlink{0000-0001-8521-1293}\thanks{Corresponding Author} \and
% Garvita Tiwari\inst{1,2}\orcidlink{0000-0002-3484-942X} \\
% Gerard Pons-Moll\inst{1,2,3}\orcidlink{0000-0001-5115-7794}}

\author{Margaret Kostyrko\textsuperscript{1,2,*}\orcidlink{0009-0004-2695-4897}, 
Yuxuan Xue\textsuperscript{1,2,*,\Envelope}\orcidlink{0000-0001-8521-1293}, 
Garvita Tiwari\textsuperscript{1,2}\orcidlink{0000-0002-3484-942X}, \\
Gerard Pons-Moll\textsuperscript{1,2,3}\orcidlink{0000-0001-5115-7794}}
\authorrunning{M.~Kostyrko \etal}

\institute{\textsuperscript{1}University of T\"ubingen \quad
\textsuperscript{2}T\"ubingen AI Center \\
\textsuperscript{3}Max Planck Institute for Informatics \\[2pt]
\textsuperscript{*}Equal Contribution \quad
\textsuperscript{\Envelope}Corresponding Author \\
\url{https://yuxuan-xue.com/avaimg}}
% They state to use \Envelope for Corresponding Author 

\maketitle

\begin{abstract}

The representation of 3D clothed humans as standardized 2D UV texture and displacement maps over an underlying body model has long been studied. 
This compact representation is enticing as it enables pretrained image networks to process, generate, and edit 3D avatars, but is only useful if 
% To be useful, 
% however, scans must be accurately aligned and brought into correspondence via high-fidelity registration. 
scans are accurately aligned and brought into correspondence via high-fidelity registration.
This prerequisite has never been met, which we argue explains the limited quality of prior UV-based methods for clothed humans. Despite its significance, no public method produces high-fidelity \smplx\smpld\ registrations with UV texture from arbitrary clothed scans.
We present \avaimg, a multi-stage optimization pipeline, to close this
% registration
gap: it enforces body-inside-clothing constraint via signed winding numbers, made viable by a three-level efficiency cascade ($\sim$10$\times$ runtime reduced, $\sim$95\% storage saved), and recovers fine surface detail using coarse-to-fine displacement optimization.
% Evaluations across 
% six datasets show that 
\avaimg\ outperforms all baselines in body fitting, shape estimation, and surface registration  across six datasets, yielding textured registrations near-indistinguishable from scans $(\text{PSNR}{=}34.48 \text{dB})$. 
% To substantiate the Avatar-as-Image representation as imminently ready for consume by image foundation models, we auto-encode our UV maps with the frozen VAE from FLUX, 
% achieving only $0.76 \text{mm}$ added Chamfer error relative to scan---and indicating 
% % This shows 
% % that the 
% that avatar-images lie within the model’s distribution, which ultimately conceptually facilitates the use of 2D generative priors for 3D avatar generation.
For validation of \avaimg's Avatar-as-Image representation as imminently compatible with image foundation models, we auto-encode our UV maps via the frozen FLUX VAE.
This achieves only $0.76 \text{mm}$ added Chamfer error relative to scan and shows that 
the resulting maps lie within natural-image distributions, supporting the use of 2D generative priors for 3D avatar generation.
Code, data, and Singularity containers will be publicly released.
\end{abstract}

\section{Introduction}
\label{sec:intro}
% \YX{Teaser: handle incomplete scan, better fitting, UV representations, editing}
% \GPM{SMPL learns linear model and vertex displacement; If you want to leverage recent modern advantages such as 2D generative prior and for 3D modeling, it will be very useful to be able to translate 3D cloth shape into images; People tried in the past but they are limited by registration quality and hence the generation is really poor; What we need to revisit this concept is a super strong registration tool, which flawlessly can represent 3D avatar as images}

\begin{figure*}[t]
  \centering
  \includegraphics[width=\linewidth]{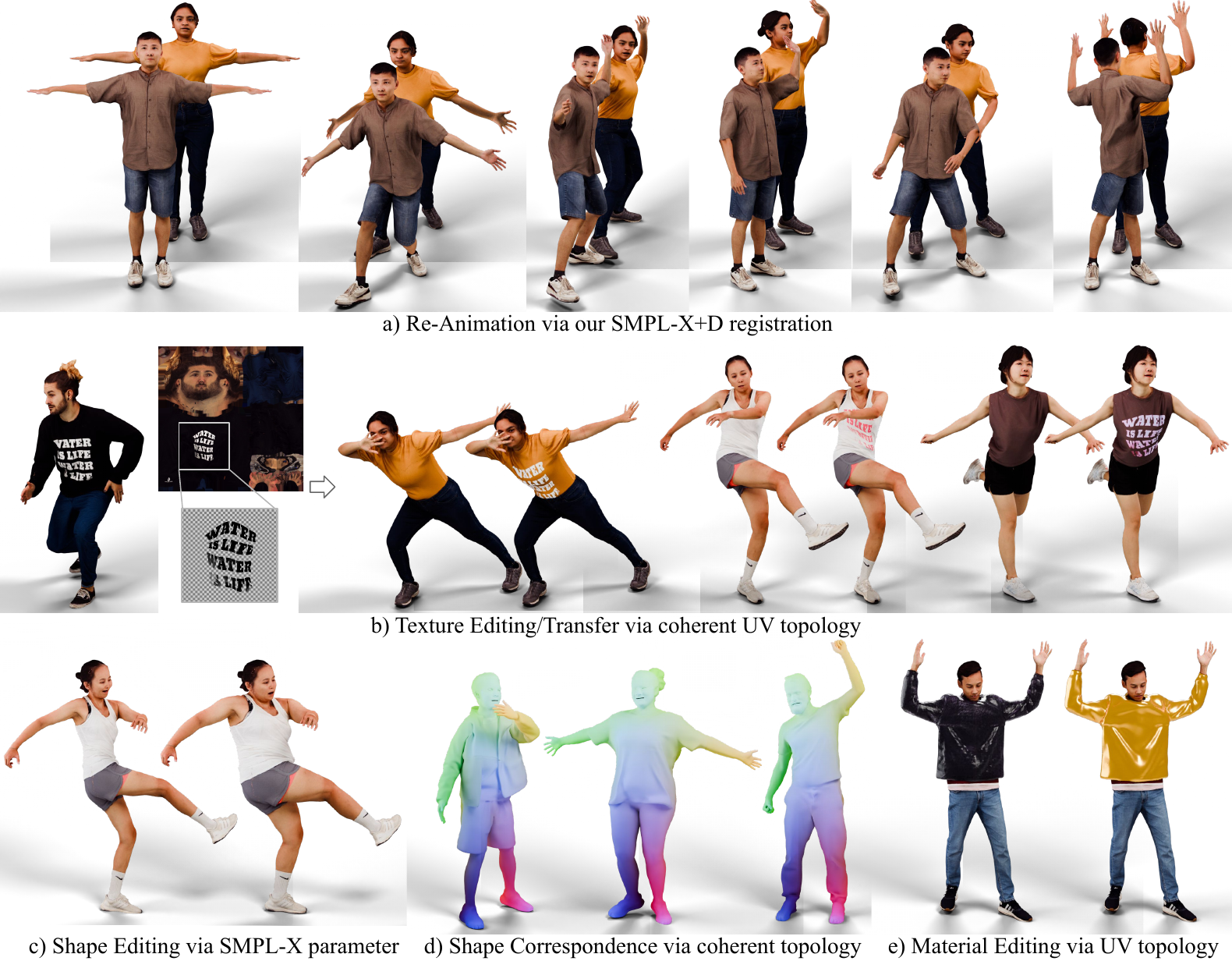}
  % [OLD — replaced by Claude 2026-03-04 (compressed caption)]
  % \textbf{Numerous applications enabled by \avaimg.}
  % \textbf{(a)}~Re-animation via \smplx\ skinning: ... (14 lines → 4)
  \caption{%
    \textbf{Applications enabled by \avaimg.}
    Because all registrations share \smplx\ parameter space and a coherent UV layout, our output directly supports re-animation, texture editing
    and transfer, shape editing, dense correspondence, and material editing.
  }
  \vspace{-2em}
  \label{fig:applications}
\end{figure*}

% Generative Clothed Human Modelling
High-fidelity digital doubles of clothed humans are integral to virtual reality, gaming, telepresence, and generative content creation.
Parametric human body models like \smplx~\cite{Loper2015SMPL, Pavlakos2019SMPLX} have long been the backbone of digital human modeling.
Historically, learning these body and clothing models relied on high-quality registration pipelines,
% were the driving force behind learning such body models and clothing models 
pertaining from SMPL~\cite{Loper2015SMPL} and Dyna~\cite{Pons-Moll2015Dyna} to ClothCap~\cite{Pons-Moll2017ClothCap} and BuFF~\cite{Zhang2017BUFF}.
Concurrently, 2D generative AI---particularly large-scale diffusion models~\cite{BlackForestLabs2024FLUX}---has produced powerful visual priors per training on billions of images.
Applying these 2D priors to 3D clothed human generation~\cite{Tang2025HGD}, texture synthesis~\cite{Sanyal2024SCULPT,Kim2024PaintIt}, and editing is thus a central goal of the field today.

% 2D representation for 3D avatar: not new but appealing; It has own problem and hence not working well
To bridge the gap between 3D geometry and 2D generative models, a natural and established approach is to represent 3D clothed humans as standard 2D images. 
Since parametric body models such as \smplx\ share UV layout across instances, the geometry and appearance of a dressed human can be coherently unrolled onto UV texture and displacement maps. These define standardized images where the same pixel always corresponds to the same anatomical location, non-pertaining to subject.
This \emph{Avatar-As-Image} concept is appealing: off-the-shelf image networks and diffusion models can utilize it to directly process, edit, and generate 3D humans without specialized 3D architectures~\cite{Alldieck2019Tex2Shape, Habermann2021DDC, Tang2025HGD}.
However, previous attempts at UV-based clothed human modeling have yielded poor results, as the prerequisite for quality was never met: the fidelity of UV maps was limited by the accuracy of the underlying 3D registration. 
When fitted mesh deviates from scan surface, the unwrapped UV images inherit every geometric and texture artifact; generative models trained on such data cannot learn physically meaningful priors.

% Recent status on registration: has their problem in accuracy; Learning-based method face chicken-and-egg problem
% Yet today, 
Concurrently, robust public frameworks capable of registering clothed scans to a unified topology with UV texture are largely unavailable. The only publicly available tool is RVH-Mesh-Registration~(RMR)~\cite{Bhatnagar2020RMR, Bhatnagar2020IPNet, Bhatnagar2020LoopReg}, which supports neither SMPL$-$X nor UV texture mapping, and uses an unsigned distance metric which cannot  distinguish whether the body approaches the scan from the in-
or outside.
Furthermore, the ``ground-truth'' registrations shipped with established datasets (BuFF~\cite{Zhang2017BUFF}, THuman~\cite{Yu2021Function4D}, and 2K2K~\cite{Han20232K2K}) demonstrate pervasive body-clothing interpenetration.
Irrespective of the underlying cause, be it unsigned distances, unreliable surface normals, or otherwise insufficient containment heuristics, the observable result is physically corrupt reference data.
Learning-based registration methods~\cite{Feng2023ArtEq,Li2025ETCH,Marin2024NICP, li2026etchx, wang2026dirtymocap, cai2026omnifit} could in principle address this at scale, but they themselves require high-quality, accurate registrations for training, constituting a chicken-and-egg problem only optimization-based pipelines can break.

% How we step in? 
We present \avaimg, a multi-stage optimization pipeline which closes this registration gap 
% in manner such that 
to make the established Avatar-As-Image
% such that the 
% and thus makes the
% Avatar-As-Image is made 
concept practically viable. 
Contrary to prior pipelines relying on surface normals to resolve body-clothing sidedness~\cite{Pons-Moll2017ClothCap,Zhang2017BUFF}, \avaimg\ enforces body-inside-clothing containment via signed winding numbers~\cite{Jacobson2013Winding} that facilitate volumetric inside/outside testing robust to noisy or open scan surfaces---regions where normals may often fail.
A three-level efficiency cascade (scan decimation, precomputed voxel grids, winding band reduction) reduces runtime by ${\sim}10\times$ and storage by ${\sim}95\%$; enabling applicability at dataset scale.
% making this physics constraint practical at dataset scale.
Robust multi-view 3D keypoint lifting
via multi-layered pose-outlier suppression
provides reliable pose initialization
% pertinent also 
across also non-standard configurations, and a coarse-to-fine displacement formulation with fourth-power edge coupling at high resolution recovers sub-centimeter geometric detail, which frequently surpasses fidelity of dataset ground-truth.
Our pipeline handles real-world scan imperfections (noise, missing regions, broken extremities) based on \smplx\ manifold prior and adaptive region-based constraints, and operates reliably across datasets of widely varying scan quality. 
Our scan-faithful registration achieves a clean Avatar-As-Image representation viable for image generative models, for which we validate immediate compatibility by encoding and decoding \avaimg's UV maps through the frozen VAE of FLUX~\cite{BlackForestLabs2024FLUX}, which accumulates only $0.76\text{mm}$ Chamfer error over scan without retraining---thus confirming the maps lie within natural-image distributions.
% % As AAI are in repr of models, we shows ours too ... 
% % In consequence to our precise 3D registration, resulting UV maps are adequately clean for modern image generative models: 
% % encoding and decoding our UV maps through the frozen VAE of FLUX~\cite{BlackForestLabs2024FLUX} accumulates only $0.76\text{mm}$ Chamfer error over scan without retraining
% % ---which confirms that registration quality is what makes the Avatar-As-Image concept serviceable.
% We validate the compatibility of Avatar-As-Image representation with 
% natural-image distributions of image generative models via 
% encoding and decoding our UV maps through the frozen VAE of FLUX~\cite{BlackForestLabs2024FLUX}, which accumulates only $0.76\text{mm}$ Chamfer error over scan without retraining.
% % ---which confirms that registration quality is what makes the Avatar-As-Image concept serviceable.
% Due to precise 3D registration, our resulting UV maps are adequately clean for various further application. 
\avaimg's coherent topology enables examples of texture transfer, appearance editing and reanimation, as showcased in \Cref{fig:applications}.
The entire pipeline will be released as a self-contained package via Singularity containers for end-to-end deployment without need for manual dependency management. In summary, our contributions are:
\begin{itemize}
    \item \textbf{\avaimg}, a multi-stage optimization pipeline for high-fidelity
    \smplx\smpld\ registration with UV texture mapping,
    supporting arbitrary clothed human scans, including noisy and incomplete
    real-world captures. Code, data, and Singularity containers will be publicly
    released.

    \item \textbf{Efficient} \textbf{signed} \textbf{winding} \textbf{numbers} for physics-aware body fitting,
    replacing surface-normal sidedness heuristics of prior works with a
    robust volumetric containment constraint, made practical 
    % making robust registration practical 
    at dataset scale
    via a three-level efficiency cascade that reduces runtime by
    ${\sim}10\times$ and storage by ${\sim}95\%$.

    \item 
    % \textbf{A Revisit of Avatar-As-Image representation}, where we show that
    % UV texture and displacement maps produced at \avaimg's registration fidelity
    % successfully encode and decode through a frozen image diffusion VAE, ultimately confirming
    % that high-quality registration is what this long-explored concept has required.
    A validation for direct compatibility of \avaimg's attained Avatar-As-Image representation with off-the-shelf image generative models, shown by successful encode-decode through a frozen diffusion VAE,
    confirming its UV texture and displacement maps reside within natural-image distributions.
    % via successful encode and decode through a frozen diffusion VAE.
\end{itemize}

\section{Related Work}
\label{sec:related}

\subsection{Clothed Human Registration}
\label{sec:relatedclothed}

\begin{wraptable}{r}{0.45\linewidth}
    \centering
    \footnotesize
    \renewcommand{\arraystretch}{1.}
    \renewcommand{\tabcolsep}{2.5pt}
    \vspace{-23pt}
    \resizebox{\linewidth}{!}{
    \begin{tabular}{@{}l ccc@{}}
        \toprule
        \multirow{2}{*}{\textbf{Method}} & \multirow{2}{*}{\begin{tabular}{@{}c@{}}\textbf{Body} \\ \textbf{Fitting}\end{tabular}} & \multirow{2}{*}{\begin{tabular}{@{}c@{}}\textbf{Surface} \\ \textbf{Registration}\end{tabular}} & \multirow{2}{*}{\begin{tabular}{@{}c@{}}\textbf{Texture} \\ \textbf{Registration}\end{tabular}} \\
        & & & \\
        \midrule
        Arteq~\cite{Feng2023ArtEq} & \cmark & \xmark & \xmark \\
        Etch~\cite{Li2025ETCH} & \cmark & \xmark & \xmark \\
        IP-Net~\cite{Bhatnagar2020IPNet} & \cmark & \cmark & \xmark \\
        LoopReg~\cite{Bhatnagar2020LoopReg} & \cmark & \cmark & \xmark \\
        PTF~\cite{Wang2021PTF} & \cmark & \cmark & \xmark \\
        \midrule
        \textbf{Ours} & \cmark & \cmark & \cmark \\
        \bottomrule
    \end{tabular}}
    \vspace{-8pt}
    \caption{
    % \avaimg~is the only usable method for body fitting, surface registration, and texture registration.
    \avaimg~introduces conjoined capacity for body fitting, surface registration, and texture mapping.
    }
    \label{tab:methods-comparison}
    \vspace{-2.0em}
\end{wraptable}

Registration of a parametric body model to clothed 3D scans revolves around a coupled pair of
objectives: naked body fitting beneath clothing, and the capture of outer clothing
surface in form of per-vertex displacement (\smplx\smpld).

\noindent\textbf{Body Fitting.}
SMPLify~\cite{Bogo2016SMPLify} and SMPLify-X~\cite{Pavlakos2019SMPLX} both fit body
parameters from 2D joint detections via optimization, but are sensitive towards
initialization.
Learning-based alternatives (LVD~\cite{Corona2022LVD},
ArtEq~\cite{Feng2023ArtEq}, ETCH~\cite{Li2025ETCH}) are faster and more
robust to pose variation, however, remain \emph{body-centric}: they recover the naked
body skeleton and shape without capturing clothing geometry or texture.

\noindent\textbf{Clothed Surface Registration.}
ClothCap~\cite{Pons-Moll2017ClothCap} and BuFF~\cite{Zhang2017BUFF} register
SMPL $+$D to 4D scan sequences using surface normals or temporal fusion to resolve
body-clothing sidedness, but both require multi-frame input and are not publicly
available.
RVH-Mesh-Registration~(RMR)~\cite{Bhatnagar2020RMR}, the only public tool, uses
unsigned distances, produces no texture, and is limited to SMPL$+$H.
Learning-based methods (IPNet~\cite{Bhatnagar2020IPNet},
PTF~\cite{Wang2021PTF}, NICP~\cite{Marin2024NICP},
LoopReg~\cite{Bhatnagar2020LoopReg}) can produce SMPL$+$D registrations, yet all
require ground-truth registrations for training, which creates a chicken-and-egg
problem when those registrations themselves contain 
% body-clothing penetration
artifacts.
No publicly available pipeline currently yields high-fidelity \smplx\smpld\
registrations with UV texture mapping from arbitrary single-frame clothed
scans; \avaimg\ completes this gap.

\subsection{2D Representations for 3D Humans}

Representation of clothed 3D humans in UV space of parametric body models has
been well explored.
% explored extensively; 
Early work showed SMPL UV maps can capture both texture and
geometry from images or
video~\cite{Alldieck2018VideoAvatar,Alldieck2018DetailedAvatars,Alldieck2019Tex2Shape,Lazova2019360Tex,Lahner2018DeepWrinkles,Mir2020Pix2Surf},
framing shape regression as 
image-to-image translation in UV
space~\cite{Alldieck2019Tex2Shape} or predicting full $360\degree$ textures from
partial observations~\cite{Lazova2019360Tex}.
These same UV maps are presently standard
% The same UV parameterization has since become standard
for neural
rendering and dynamic character animation~\cite{Habermann2021DDC,Liu2021NeuralActor,Pang2024ASH,Zhu2024TriHuman,Sun2025DUT,Xue2023NSF},
where motion-dependent appearance is generated directly in texture space.
Recent UV-space methods include
% has been targeted by
SMPLitex~\cite{Casas2023SMPLitex}, Paint-It~\cite{Kim2024PaintIt},
SCULPT~\cite{Sanyal2024SCULPT} and
Chaudhuri~\etal~\cite{Chaudhuri2021SemiSupTex}, while encoding 3D geometry as
2D images to leverage diffusion priors is an emerging trend both for general
objects~\cite{Yan2025Omages,Xue2026GeoRelight} and clothed
humans~\cite{Tang2025HGD,Xue2024Human3Diffusion,Xue2025Gen3Diffusion,Xue2025InfiniHuman}.

All methods above are limited by registration quality:
without high-fidelity \smplx\smpld\ meshes, resulting UV maps inherit
geometric artifacts and texture misalignment.
\avaimg\ addresses this bottleneck by producing registrations whose UV
maps can be imminently consumed by pretrained image
VAEs~\cite{BlackForestLabs2024FLUX} and diffusion models without modification.

\section{Method}
\label{sec:method}

\begin{figure*}[t]
  \centering
  \includegraphics[width=\linewidth]{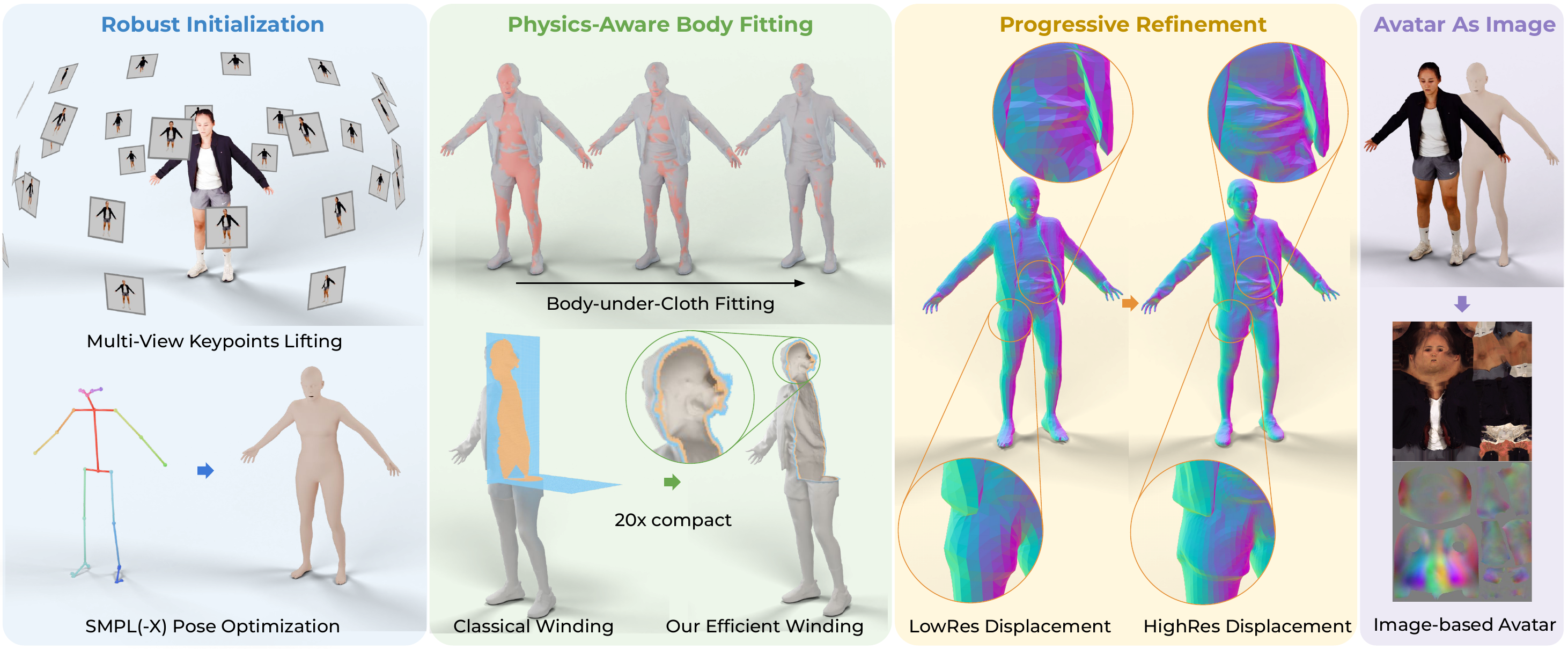}
  % [OLD — replaced by Claude 2026-03-04 (compressed caption)]
  % \textbf{Method Overview.} Given a clothed 3D scan, \avaimg\ proceeds in four stages.
  % \textbf{(1)~Robust Initialization}: ... (7 lines → 4)
  \caption{%
    \textbf{Method Overview.}
    Given a clothed 3D scan, \avaimg\ proceeds in four stages:
    (1)~multi-view keypoint lifting for robust pose initialization,
    (2)~physics-aware body fitting via efficient signed winding numbers,
    (3)~coarse-to-fine displacement optimization for fine surface detail, and
    (4)~remapping into 2D texture and displacement maps.
  }
  \label{fig:pipeline}
  \vspace{-2em}
\end{figure*}

\edit{Given only a raw 3D scan $\mathcal{S} = \{v, f, v_t\}$ with vertices $v \in \mathbb{R}^{N_{3D}\times 3}$, faces $f \in \mathbb{N}^{M\times 3}$, and UV coordinates $v_t\in \mathbb{R}^{N_{2D}\times 2}$}{Given only a raw 3D scan $\mathcal{S} = \{v, f, v_t\}$ with vertices $v$, faces $f$, and UV coordinates $v_t$}, \avaimg\ produces a \smplx\smpld\ registration $M(\gamma, \beta, \theta, \psi, D)$ paired with UV texture and displacement maps that jointly define the Avatar-As-Image representation.
Three coupled challenges must be resolved for high-fidelity registration: \edit{estimating pose under clothing occlusion, fitting the naked body beneath clothing without direct observation, and capturing the complex clothing surface geometry.}{robust pose initialization, naked body fitting below occluding clothing, and complex clothing surface capture.}

Two design principles guide every stage of our pipeline.
\emph{Coarse-to-fine Optimization}: we constrain heavily at earlier stages to steer away from local minima, then progressively relax constraints to recover fine detail.
% This principle manifests in staged pose parameter unfreezing (Sec.~\ref{sec:pose_init}), body prior annealing (Sec.~\ref{sec:body_fitting}), mesh resolution progression (Sec.~\ref{sec:surface_reg}), and loss weight scheduling throughout.
\edit{(2) \emph{adaptive robustness}: all distance computations use Geman-McClure robustification to downweight outliers, region-dependent weights adapt to observed data quality (\eg automatically constraining unobserved hands), and the \smplx\ topology itself acts as a smooth manifold prior that regularizes noisy or incomplete scans.}{\emph{Innate Robustness}: we leverage the \smplx\ topology itself as a smooth, complete manifold prior, which grounds regularization across scan noise, lacking geometry and merged self-contact regions to produce clean, consistent meshes.}
% \smplx\ topology provides a complete, smooth manifold prior: even when scans contain holes, broken hands, or noisy geometry in self-contact regions (\eg armpits, crotch), the registration produces clean, complete meshes

% ---------------------------------------------------------------------------
\subsection{Robust Pose Initialization}
\label{sec:pose_init}
% ---------------------------------------------------------------------------

\noindent
Pose estimation constitutes our pipeline's foundation: errors here onwards propagate irrecoverably across all subsequent stages.
Initializing via mean pose, as in prior work~\cite{Bhatnagar2019MGN}, routinely traps the optimization in local minima on non-standard configurations, such as raised arms or crossed legs.
Instead, we lift reliable 3D joint targets from the scan itself, providing scan-specific initialization.

\noindent\textbf{Scan Normalization.}
To handle heterogeneous datasets from widely varying coordinate frames, \edit{we first normalize each scan to a $[-1, 1]^3$ bounding box, then align it to \smplx\ dimensions via a height-based scale factor and centroid translation derived from gender-specific body statistics.}{we normalize each scan to gendered \smplx\ via a height-based scale factor and centroid translation.}
% This two-step normalization ensures that scans from datasets as different as Thuman~\cite{Yu2021Function4D} (in meters) and BuFF~\cite{Zhang2017BUFF} (in millimeters) enter the pipeline in a consistent coordinate frame. 
% > > > MK{No, during OpenPose we norm to -1,1 because that is independent of user specified SMPL type / gender, we later norm to SMPL during registration as only then do the flags play a role}

\noindent\textbf{Multi-View Keypoint Lifting.}
\edit{We render $V = 72$ views of each textured scan using PyTorch3D~\cite{Ravi2020PyTorch3D} at three elevation levels with full azimuth coverage.
OpenPose~\cite{Cao2021OpenPose} detects $K = 137$ keypoints (body, hands, face) per view, each with a confidence score $c \in [0, 1]$.
Multi-view bundle adjustment triangulates these detections into 3D joint locations $J^{3D} \in \mathbb{R}^{K \times 3}$ by minimizing:
\begin{equation}
    \mathcal{L}_{\text{kp}} = \frac{1}{V} \sum^V_{v=1}\sum^K_{k=1} c_{v,k}^2 \, \mathbf{1}(c_{v,k} \geq 0.3) \; \| J^{2D}_{v,k} - P_v(J^{3D}_k) \|_1,
    \label{eq:keypoint}
\end{equation}
where $P_v(\cdot)$ projects to image coordinates.
Three layers of outlier suppression work together: the L1 norm is inherently robust to large residuals, squaring the confidence scores aggressively downweights uncertain detections, and the hard threshold at $0.3$ zeros out unreliable keypoints entirely, \eg those hallucinated outside the image frame. To eliminate any inconveniency due to OpenPose runtime compilation, we provide end-to-end GPU-friendly singularity container for user-friendly deployment of our \avaimg.}{Per textured scan, we render $72$ views across $3$ elevation levels with full azimuth coverage using PyTorch3D~\cite{Ravi2020PyTorch3D}.
OpenPose~\cite{Cao2021OpenPose} is employed to detect $137$ keypoints 
% (body, hands, face) 
with respective confidence scores per view.
% , each with a confidence score.
We perform multi-view bundle adjustment to triangulate detected keypoints into 3D joint locations, during which we apply 
% $J^{3D} \in \mathbb{R}^{K \times 3}$ by minimizing:
% \begin{equation}
%     \mathcal{L}_{\text{kp}} = \frac{1}{V} \sum^V_{v=1}\sum^K_{k=1} c_{v,k}^2 \, \mathbf{1}(c_{v,k} \geq 0.3) \; \| J^{2D}_{v,k} - P_v(J^{3D}_k) \|_1,
%     \label{eq:keypoint}
% \end{equation}
% where $P_v(\cdot)$ projects to image coordinates.
three layers of outlier suppression: the outlier robust $L1$-norm,
% is inherently robust to large residuals, 
squaring of the confidence scores, 
% to aggressively downweight uncertain detections
and a hard threshold of $0.3$ to zero out unreliable keypoints entirely}. 
To eliminate any inconveniency due to OpenPose runtime compilation, we will provide an end-to-end GPU-friendly Singularity container for user-friendly deployment of our \avaimg\ pipeline.
% > > > \MK{I think the joint equation should not be highlighted if we barely highlight any other equations actually pertaining to the core registration itself; this is an established equation either way}

\noindent\textbf{\smplx\ Pose Optimization.}
The triangulated targets form a joint-fitting loss $\mathcal{L}_j$, which is minimized alongside a shape prior $\mathcal{L}_\beta$ and a pose prior $\mathcal{L}_\theta$~\cite{Pavlakos2019SMPLX}.
Following the coarse-to-fine principle, we stage the optimization in three phases: (1)~global orientation, head, shoulders, and right foot; (2)~all body joints;
% with the pose prior annealed from strong ($w{=}0.06$) to moderate ($w{=}0.01$); 
(3)~full \smplx\ including hands and facial expression.
This schedule yields a robust pose that serves as reliable initialization for body fitting.

% ---------------------------------------------------------------------------
\subsection{Physics-Aware Fitting with Efficient Winding Numbers}
\label{sec:body_fitting}
% ---------------------------------------------------------------------------

\noindent
With pose established, we optimize all \smplx\ body parameters 
% $(\gamma, \beta, \theta, \psi)$ 
to situate the naked body beneath clothing; a task complicated by absence of direct observation.
% [OLD — replaced by Claude 2026-03-02 (soften unsigned claim, fix typo)]
% Minimizing unsigned distance between the \smplx\ surface and the scan drives the body \emph{toward} the clothing surface regardless of direction, regularly pushing it through the clothing.
% This is precisely the artifact present in existing dataset ground truth~\cite{Zheng2019DeepHuman,Han20232K2K,Ho2023CustomHumans} and puclicly-available registration tools, corrupting any downstream processing that assumes the body is enclosed by the clothing. \YX{I don't know if it's correct to say those related works are unsigned, need double check}
Minimizing raw distance between the \smplx\ and scan surface commonly finds solution penetrating the clothing outwards as mean to near topology.
% whether unsigned or signed via unreliable surface normals, 
% regularly pushes the body through the clothing.
This is precisely the artifact observed in existing dataset ground truth~\cite{Zheng2019DeepHuman,Han20232K2K,Ho2023CustomHumans} and publicly available registration tools.
% corrupting any downstream task that assumes the body is enclosed by clothing.

\noindent\textbf{Inside-Outside Constraint via Signed Winding Numbers.}
We exploit the physical fact that a body is always enclosed by its clothing.
Generalized winding numbers~\cite{Jacobson2013Winding} are calculated on a volumetric basis, contrary to surface-normal or ray-casting approaches, which are vulnerable to noisy or non-watertight meshes. This sensitivity may result in erroneous labeling of internal points as external and thus cause body collapse or otherwise capricious behavior during fitting. To mitigate pipeline robustness to cases of open topology, we apply winding numbers as method of sign computation, where
query points are classified relative to static scan mesh as inside ($+1$) or outside ($-1$) scan surface.
% computed on the static scan mesh classify query points as inside ($+1$) or outside ($-1$) scan surface.
% [OLD — replaced by Claude 2026-03-02 (swap rho/pReLU order, precise asymmetric description)]
% We incorporate this sign into the mesh-to-scan (m2s) distance and apply a parametric ReLU:
% \begin{equation}
%   \mathcal{L}_d = \mathrm{pReLU}\!\left[\,\rho\!\left(\mathrm{dist}_{\text{signed-m2s}}\!\left(\mathcal{S},\, M(\gamma,\beta,\theta)\right)\right)\right],
%   \label{eq:winding}
% \end{equation}
% where $\rho$ is the Geman-McClure robust function.
% The pReLU activates only for \emph{negative} signed distances---body vertices that have escaped outside the clothing---while imposing no penalty when the body approaches the scan from within.
% The body is thus free to fit tightly under the clothing, yet any penetration incurs a steep, differentiable penalty.
We incorporate this sign into the mesh-to-scan (m2s) distance loss and apply a parametric ReLU:
% with asymmetric weighting:
\begin{equation}
  \mathcal{L}_d = \rho\!\left(\mathrm{pReLU}\!\left[\mathrm{dist}_{\text{signed-m2s}}\!\left(\mathcal{S},\, M(\gamma,\beta,\theta, \psi)\right)\right]\right),
  \label{eq:winding}
\end{equation}
where $\rho$ is the Geman-McClure robust function.
The pReLU applies asymmetric weighting to the signed distances: body vertices inside of scan clothing receive a mild penalty, while vertices that have escaped outside incur an amplified price,
% by a factor of $\alpha{=}40$, 
producing a steep, differentiable barrier against penetration, only firmly enforceable due to reliable sign classification.
% without zeroing out the inside signal entirely.
% Region-dependent vertex weights tighten the constraint at the face and hands, where the body and clothing surfaces are in close contact.
The full objective retains $\mathcal{L}_j$, $\mathcal{L}_\beta$, and $\mathcal{L}_\theta$, 
% to preserve pose plausibility 
with prior weight annealed over time to gradually free body from constraint.
% with the prior weight annealed from $0.06$ to $0.03$ across two optimization loops, 
% gradually freeing the body from the prior as the data term takes over.

\noindent\textbf{Efficient Winding Computation.}
Na\"ive compute of generalized winding numbers is $\mathcal{O}(n_{\text{query}} \times n_{\text{faces}})$, prohibitive on high-polygon scans such as 2K2K~\cite{Han20232K2K} ($100$k--$200$k faces).
We introduce a three-level efficiency cascade that makes the physics constraint practical at dataset scale.
\emph{First}, 
the scan mesh is decimated to approximately $10\%$ of its original face count (min. $40$k) for winding computation. 
% using mesh collapse; 
% only this lightweight proxy is used for winding computation, while all fitting stages operate on the full-resolution scan.
\emph{Second}, winding numbers are precomputed on a dense voxel grid (5\,mm) around the decimation bounding box and are thresholded to a binary ($\pm 1$) field.
During optimization, classifying a body vertex thus requires only a nearest-neighbor look-up. 
% no winding recomputation is needed per iteration.
\emph{Third}, we apply \emph{winding band reduction}: only boundary voxels (those whose $7$ nearest neighbors include a sign change) 
% plus a one-hop propagation shell 
are retained; the remaining ${\sim}95\%$ of the voxel grid is discarded, such that both storage and query time are reduced by roughly an order of magnitude.
% ; viable as the body surface lies near this inside-outside boundary and queries against distant, sign-uniform voxels are unnecessary.

% \noindent\textbf{Adaptive Hand Constraints.}
% When the maximum keypoint confidence for a hand falls below $0.2$, we treat it as unobserved: the coupling weight for hand vertices is raised to $1000$ and the pose prior for hand joints to $10^5$, effectively locking the hand to the \smplx\ template.
% This prevents unobserved regions (where scans commonly exhibit noise or missing geometry) from degrading the fit as in~\cref{fig:fid_protocol} (a). 

% ---------------------------------------------------------------------------
\subsection{Progressive Refinement: From Coarse Body to Fine Detail}
\label{sec:surface_reg}
% ---------------------------------------------------------------------------

\noindent
Body fitting recovers the underlying body shape but not exterior scan surface: geometric structure, folds, wrinkles, and hair that define visual appearance.
We capture these via per-vertex displacements $D$ in a two-pass coarse-to-fine scheme, embodying the progressive refinement principle at mesh resolution level.
% , loss weighting, and constraint scheduling.

\noindent\textbf{Low-Resolution Displacement.}
With the body now correctly positioned inside the scan and not our focus anymore, the penetration constraint is no longer needed.
We switch to unsigned data terms and optimize free vertex positions $v_{\text{free}}$ stemming from the \smplx\ body mesh, and introduce further loss terms for optimization.
% by minimizing:
% \begin{equation}
%   \mathcal{L}_{\text{surf}} = w_d\mathcal{L}_d + w_j\mathcal{L}_j + w_u\mathcal{L}_u + w_c\mathcal{L}_c + w_l\mathcal{L}_l + w_\beta\mathcal{L}_\beta + w_\theta\mathcal{L}_\theta.
%   \label{eq:surf}
% \end{equation}
$\mathcal{L}_u$ penalizes the difference between displacements in posed and canonical (unposed) space, promoting topology independence from articulation.
$\mathcal{L}_c$ is an edge-coupling loss that penalizes edge-length deviations from the non-displaced body, with weighting assigned per select region of bone-based skinning weights of \smplx, restricting deformation in specified areas (\eg hands). 
% : skeletal regions (elbows, shoulders, hips) receive high coupling weights that preserve body structure, while clothing regions deform freely.
$\mathcal{L}_l$ applies cotangent Laplacian smoothing to suppress displacement noise, where weights are assigned by employ of designed vertex maps. \emph{Scheduled weight annealing} 
% This pass runs in four sub-phases with \emph{scheduled weight annealing}: the edge-coupling weight decreases from $2.0$ to $0.4$ and the Laplacian weight from $25$ to $4$ as the optimization progresses, 
allows the mesh to initially capture the global silhouette under stronger regularization, then relaxes to fit finer deforms.
% Head region weights are also adapted per phase ($6 \to 4 \to 5 \to 1$), accommodating the transition from coarse head shape to detailed hair geometry.

\noindent\textbf{High-Resolution Displacement.}
The low-resolution displaced mesh is smooth subdivided via Loop subdivision~\cite{loop1987smooth}, approximately quadrupling the face count.
Now on high-resolution mesh, a second displacement pass parallel to the one prior recovers fine surface detail via three targeted modifications. 

\emph{First}, the data term switches to unsigned s2m only, directing the mesh toward the outer scan clothing surface without bidirectional pull, where the data adaptive multiplier progressively increases and tightens scan adherence. 
% An adaptive data weight multiplier increases from $1.0$ to $1.5$ across the optimization, progressively tightening scan adherence (coarse-to-fine applied within a single pass).
% [OLD — replaced by Claude 2026-03-02 (punchier CPL^4 contrast)]
% Second, the edge-coupling term is raised to the fourth power---$(w_c\mathcal{L}_c)^4$.
% At low resolution, squared coupling allows bulk clothing deformation.
% At high resolution, the fourth power creates a dramatically steeper penalty around edge-length violations: the mesh must \emph{snap} to wrinkles and folds rather than smoothly averaging over them.
% This is the key mechanism enabling sub-centimeter geometric detail.
\emph{Second}, the edge-coupling term $\mathcal{L}_c$ is additionally squared,
% (raised to fourth power under squared L2), 
% resulting in total fourth power under squared L2-Loss.
% Unlike standard squared coupling, which over-smooths surface geometry by averaging over edge-length violations, 
% The fourth power 
% which creates a much steeper penalty landscape: the mesh must \emph{snap} to high-frequency wrinkles and folds instead of smoothly averaging over them.
% This is what enables sub-centimeter geometric detail.
which produces a loss at a flatter basin but steeper sidewalls: small mesh deviations \emph{snap} towards high-frequency geometry shifts (\eg folds, cloth edges, heels) while maintaining overall consistent topology.  
\emph{Third}, alterations to region-dependent weights explicitly mitigate where fine structures are perceptually critical and likely otherwise maladaptive.
Per experiments in Sections \ref{sec:exp_body_fitting}, \ref{sec:ablation}, we substantiate that our achieved level of geometric fidelity frequently surpasses that of comparable baselines,
% of the original dataset ground truth, 
and can even compensate noise and lack of scan geometric completion via leverage of \smplx\ topology as manifold prior (Sec. \ref{sec:gt_comparison})---enabling \avaimg\ to operate reliably across datasets of largely variable scan quality.

% \noindent\textbf{Robustness to Scan Imperfections.}
% Geman-McClure robustification is applied to all distance terms throughout, with tighter $\sigma$ for m2s ($0.001$) than for s2m ($0.01$), reflecting the different noise profiles of each direction.
% The \smplx\ topology provides a complete, smooth manifold prior: even when scans contain holes, broken hands, or noisy geometry in self-contact regions (\eg armpits, crotch), the registration produces clean, complete meshes as in~\cref{fig:fid_protocol} (a).
% Combined with the adaptive hand constraints from Sec.~\ref{sec:body_fitting}, this enables \avaimg\ to operate reliably across datasets of very different scan quality.

% ---------------------------------------------------------------------------
\subsection{The Avatar-As-Image Representation}
\label{sec:texture_remap}
% ---------------------------------------------------------------------------

\noindent
The preceding stages yield a high-quality \smplx\smpld\ registration with canonical UV mapping shared across all subjects.
% Here we describe how that registration becomes the 
To achieve Avatar-As-Image representation, we convert each registration into a pair of standardized images:
a UV texture and a UV displacement map per \smplx\ UV space, where the same pixel always corresponds to the same anatomical location irrespective of subject. 
This semantic alignment facilitates the maps as suitable training data for image generative models with no additional alignment or preprocessing.

\noindent\textbf{Precomputed UV Lookup Maps.}
We precompute an \emph{f-map} and a \emph{b-map} at target image resolution.
The f-map specifies which high-res \smplx\ UV face each pixel at resolution places within; and the b-map stores the barycentric coordinates of said pixel within respective face.
Both maps depend only on the \smplx\ UV topology, not on the scan, and are therefore computed once and reused for all subjects.
Changing the output resolution (\eg $512^2 \rightarrow 4096^2$) or switching the \smplx\ body model variant 
% (SMPL, SMPL-H, \smplx) 
requires only one-time computation of these lightweight lookup maps.

\noindent\textbf{Texture and Displacement Transfer.}
For each resolution pixel, the f-map and b-map are used to locate the corresponding 3D point on the high-resolution registered mesh;
% (critically, the \emph{subdivided} mesh from Sec.~\ref{sec:surface_reg}, so that fine geometric detail is captured in the UV maps).
from which we identify the nearest scan surface point, express it in barycentric coordinates within the closest scan face, and bilinearly sample the scan's color at the corresponding scan UV location.
Pixels without coverage are filled in by 
morphological 
inpainting.
Scans with vertex colors instead of textures interpolate color directly from the nearest scan vertices.
Applying the same procedure to per-vertex displacements yields the UV displacement map.

Together, UV texture and displacement maps constitute a complete Avatar-As-Image representation.
In \Cref{sec:vae_experiment}, we validate they encode-decode through the frozen VAE of a pretrained image diffusion model at high fidelity without any fine-tuning, confirming their readiness as training data for generative models.

\section{Experiments}

\begin{figure*}[t]
  \centering
  \includegraphics[width=\linewidth]{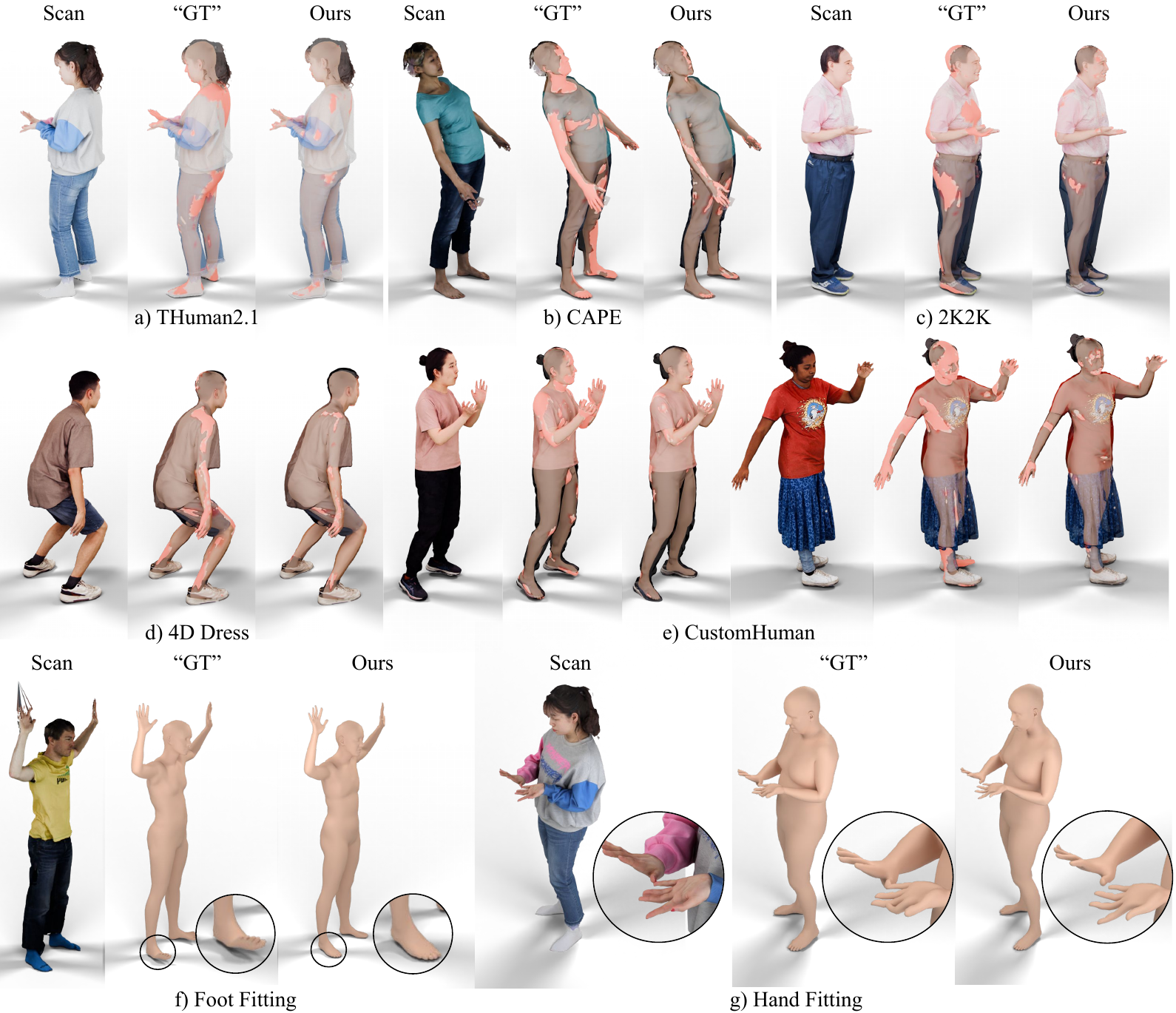}
  \caption{%
    \textbf{Comparison with dataset ground truth.}
    Body fitting overlaid on clothed scan surface across five datasets (penetrating vertices highlighted in red).
    The dataset GT fits exhibit pervasive penetration across the entire body,
    while ours keep the body enclosed within the clothing (\cref{tab:better_than_gt}).
    Zoomed insets show that dataset GT produces bent, misaligned
    feet and stiff finger articulation, while our method recovers
    anatomically plausible poses and handles noisy hand geometry.
  }
  \label{fig:gt_comparison}
    \vspace{-2em}
\end{figure*}

\subsection{Evaluation Benchmarks}
\label{sec:eval_benchmark}

We evaluate \avaimg\ across six public scan datasets spanning diverse body shapes,
poses, and clothing styles:
4D-Dress~\cite{Wang20244DDress} (47~subjects), BuFF~\cite{Zhang2017BUFF}
(26), CAPE~\cite{Ma2020CAPE} (40),
THuman2.1~\cite{Yu2021Function4D} (20), 2K2K~\cite{Han20232K2K} (20), and
CustomHuman~\cite{Ho2023CustomHumans} (20).

We assess four complementary aspects of registration quality:
\textbf{Body fitting} is evaluated on the first three datasets via \emph{penetration rate}
(\% of body vertices outside scan surface), \emph{penetration depth}
(mean distance of penetrating vertices to scan surface), and
\emph{scan proximity} (mean \smplx\ to scan distance) 
% (Sec. \ref{sec:gt_comparison}, \ref{sec:exp_body_fitting}).
% [OLD — replaced by Claude 2026-03-03]
% Proximity alone is an insufficient metric: a body that penetrates the
% clothing achieves deceptively low distance while being physically
% implausible.
Proximity alone is an insufficient metric, as a body that penetrates the
clothing achieves deceptively low distance while being physically
implausible.
The three body fitting metrics must therefore be interpreted 
jointly, where low proximity is meaningful only when met by low
penetration rate and depth.
\textbf{Shape estimation} is evaluated on BuFF---which uniquely provides
ground-truth minimal-clothing body scans---via bidirectional Chamfer
distance in T-pose after Procrustes alignment.
\textbf{Surface registration} measures the bidirectional Chamfer distance
between the registered \smplx\smpld\ mesh and the input scan.
% [OLD — replaced by Claude 2026-03-03]
% \textbf{Texture quality} is assessed via the Fr\'{e}chet Inception Distance
% (FID)~\cite{Heusel2017FID} between multiview renderings of original scans
% and textured registrations.
\textbf{Texture registration} is assessed via multiview rendering
PSNR between original scans and textured registrations; we additionally
report the Fr\'{e}chet Inception Distance (FID)~\cite{Heusel2017FID} as a
distributional similarity measure.
All quantitative results are summarized in \Cref{tab:quantitative}.
We validate \avaimg's representational UV map compatibility with latent diffusion models through a
\textbf{VAE roundtrip} experiment (\cref{sec:vae_experiment}), and
provide \textbf{ablation studies} of our key design choices
(\cref{sec:ablation}).

\subsection{Comparison with Dataset Ground Truth}
\label{sec:gt_comparison}

\begin{table}[t]
  \centering
  \renewcommand{\arraystretch}{1.}
  \renewcommand{\tabcolsep}{2.5pt}
  \resizebox{0.995\linewidth}{!}{
  \begin{tabular}{ll cccccc}
    \toprule
    & Method
      & 2K2K~\cite{Han20232K2K}
      & THuman2.1~\cite{Yu2021Function4D}
      & CustomHuman~\cite{Ho2023CustomHumans}
      & CAPE~\cite{Ma2020CAPE}
      & 4D-Dress~\cite{Wang20244DDress}
      & Overall \\    
      \midrule
    \multirow{2}{*}{Prox.\ (mm) $\downarrow$}
      & Dataset GT & 19.5{\scriptsize$\pm$3.9} & \textbf{9.2}{\scriptsize$\pm$1.7} & \textbf{9.9}{\scriptsize$\pm$2.0} & \textbf{6.8}{\scriptsize$\pm$5.4} & 12.1{\scriptsize$\pm$2.7} & 11.8{\scriptsize$\pm$4.7} \\
      & Ours       & \textbf{12.9}{\scriptsize$\pm$2.9} & 10.0{\scriptsize$\pm$1.8} & 11.3{\scriptsize$\pm$4.4} & 8.7{\scriptsize$\pm$1.0} & \textbf{12.0}{\scriptsize$\pm$2.9} & \textbf{11.4}{\scriptsize$\pm$3.2} \\
    \midrule
     \multirow{2}{*}{Pene.\ R.\ (\%) $\downarrow$}
      & Dataset GT & 33.6{\scriptsize$\pm$9.5} & 26.2{\scriptsize$\pm$4.6} & 33.3{\scriptsize$\pm$5.7} & 50.6{\scriptsize$\pm$7.7} & 20.9{\scriptsize$\pm$3.6} & 28.2{\scriptsize$\pm$11.3} \\
      & Ours       & \textbf{10.0}{\scriptsize$\pm$3.2} & \textbf{15.9}{\scriptsize$\pm$4.2} & \textbf{16.1}{\scriptsize$\pm$6.9} & \textbf{21.4}{\scriptsize$\pm$5.6} & \textbf{14.2}{\scriptsize$\pm$5.2} & \textbf{15.0}{\scriptsize$\pm$6.0} \\
    \midrule
    \multirow{2}{*}{Pene.\ D.\ (mm) $\downarrow$}
      & Dataset GT & 14.5{\scriptsize$\pm$4.5} & 5.0{\scriptsize$\pm$1.0} & 6.0{\scriptsize$\pm$1.1} & 7.4{\scriptsize$\pm$6.5} & 4.7{\scriptsize$\pm$0.8} & 6.4{\scriptsize$\pm$4.3} \\
      & Ours       & \textbf{2.1}{\scriptsize$\pm$0.4} & \textbf{3.1}{\scriptsize$\pm$0.4} & \textbf{4.2}{\scriptsize$\pm$0.8} & \textbf{3.6}{\scriptsize$\pm$0.5} & \textbf{3.4}{\scriptsize$\pm$0.4} & \textbf{3.4}{\scriptsize$\pm$0.7} \\

    \bottomrule
  \end{tabular}
  }
  \caption{%
    \textbf{Body fitting vs.\ dataset ground truth.}
    Metrics over scan proximity, penetration rate and
    penetration depth
    of dataset-provided ``ground-truth'' \smplx\ fits vs.\ ours,
    as evaluated on five public datasets.
    Our body-fits consistently achieve lower penetration metrics, indicating
    more accurate body-under-clothing estimation.
  }
  \label{tab:better_than_gt}
   \vspace{-2em}
\end{table}

Several established scan
datasets ship ``ground-truth'' \smplx~registrations
alongside their raw scans.
% These registrations were produced by older pipelines that rely on unsigned distance metrics or surface normals, neither of which can reliably distinguish whether the body approaches the scan from inside or outside, especially on open or incomplete surfaces.
We have observed that the provided body-fits exhibit systematic artifacts:
fitted body mesh frequently \emph{penetrates} the clothing
surface, hand poses are inaccurate due to weak constraints on
articulated extremities, and foot alignment is fickle, particularly
in CAPE, where open scan surfaces at feet soles provide no geometric
anchor (\cref{fig:gt_comparison}).
These are not simply minor cosmetic issues: any downstream method trained on
penetrating body-fits obtains a physically impossible prior, and shape
estimation benchmarks that evaluate against such registrations risk
rewarding methods that replicate the same artifacts.

\noindent\textbf{Improved Body Fitting.}
By enforcing strict body-inside-clothing containment via signed winding
numbers (\cref{sec:body_fitting}), \avaimg\ eliminates the penetration
artifacts present in dataset ground truth.
\Cref{tab:better_than_gt} quantifies this across five datasets: the average penetration rate drops from 28.2\% (dataset
GT) to \textbf{15.0\%} (ours), and the average penetration depth
decreases from 6.4\,mm to \textbf{3.4\,mm}, while maintaining
comparable scan proximity (11.4\,mm vs.\ 11.8\,mm).
The improvement is most striking on 2K2K, where penetration rate decreases
from 33.6\% to 10.0\%, and on CAPE, where it drops from 50.6\% to
21.4\% in spite of challenging open regions.
Our improved fittings can serve as higher-quality
training data for learning-based methods~\cite{Li2025ETCH,Marin2024NICP},
breaking the circular dependency (chicken-and-egg problem) identified in \Cref{sec:intro}.

\begin{figure*}[t]
  \centering
  \includegraphics[width=\linewidth]{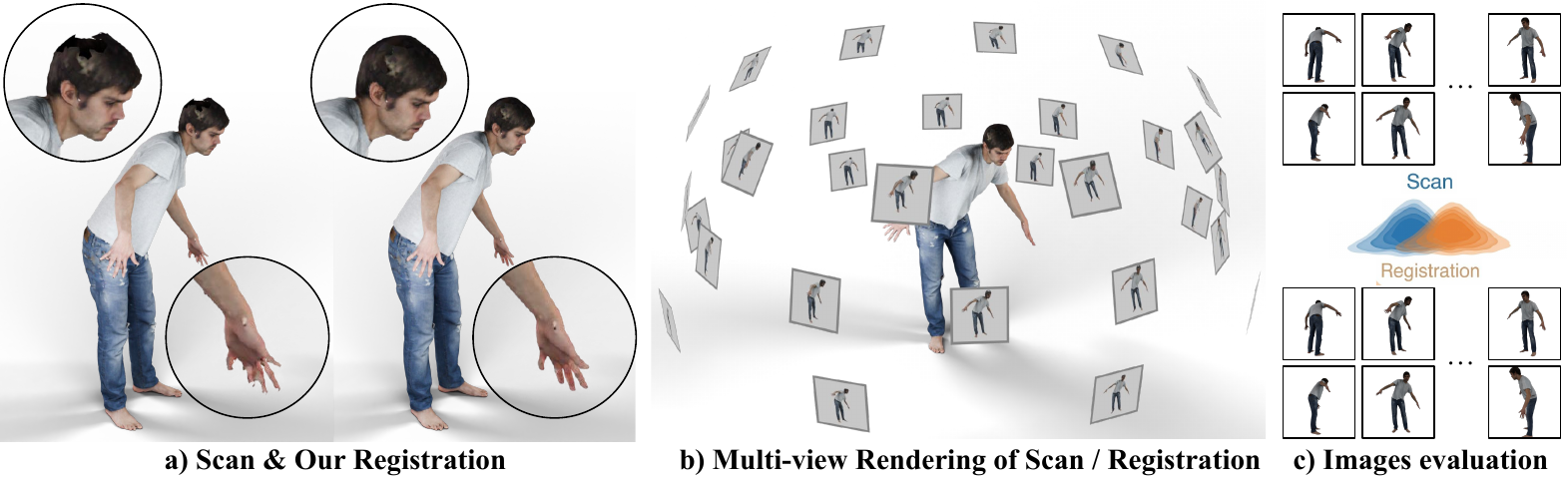}
  % [OLD — replaced by Claude 2026-03-04 (compressed caption)]
  % \textbf{(a)}~Side-by-side comparison ... (10 lines → 4)
  \caption{%
    \textbf{Texture registration and evaluation protocol.}
    Original scan (left) vs.\ our resulting textured registration (right); insets show clean recovery of noisy hand and head geometry.
    Both are rendered from multiple viewpoints and compared via image metrics ($\text{PSNR}\,{=}\,34.48$\,dB, \cref{tab:quantitative}).
  }
  \label{fig:fid_protocol}
   \vspace{-1em}
\end{figure*}

\noindent\textbf{Robustness to Scan Imperfections.}
Beyond the quality of the registrations themselves, the input scans are
frequently imperfect: \eg BuFF and THuman contain noisy hand geometry,
incomplete surface coverage, and missing body regions.
Na\"ive surface fitting propagates these defects into the output mesh, yet
\avaimg\ naturally handles such artifacts due to the \smplx\ parametric
model acting as a strong anatomical prior: the body manifold constrains
hands, faces, and extremities to plausible configurations, while the
coarse-to-fine displacement scheme (\cref{sec:surface_reg}) captures underlying
surface detail without overfitting to noise.
As a result, our registrations are often \emph{cleaner} than the input
scans in corrupted regions: the parametric prior effectively denoises
the geometry while preserving faithful surface detail elsewhere
(\cref{fig:fid_protocol}).

% [OLD — replaced by Claude 2026-03-04 (removed redundant metric definitions, already in sec 4.1)]
% \subsection{Underlying Body Fitting} ... (12-line intro + 3 metric defs removed)
\subsection{Comparison with State-of-the-Art}
\label{sec:exp_body_fitting}
\begin{table}[t]
  \centering
 \renewcommand{\arraystretch}{1.}
\renewcommand{\tabcolsep}{2.5pt}
\resizebox{0.995\linewidth}{!}{
  \begin{tabular}{l ccc c c c}
    \toprule
    & \multicolumn{3}{c}{Body Fitting}
    & Shape Est.
    & Surf.\ Reg.
    & Tex. Reg. \\
    \cmidrule(lr){2-4} \cmidrule(lr){5-5} \cmidrule(lr){6-6} \cmidrule(lr){7-7}
    Method
      & Pene.\ R.\ (\%) $\downarrow$
      & Pene.\ D.\ (mm) $\downarrow$
      & Prox.\ (mm) $\downarrow$
      & Chamfer (mm) $\downarrow$
      & Chamfer (mm) $\downarrow$
      & PSNR $\uparrow$ \\
    \midrule
    IPNet~\cite{Bhatnagar2020IPNet}  & 44.5 & 25.93 & 23.41 & 9.95            & 8.61            & --- \\
    ETCH~\cite{Li2025ETCH}           & \underline{35.0} & 13.53 & 13.01 & \underline{7.76} & ---             & --- \\
    NICP~\cite{Marin2024NICP}        & 54.6 & 12.98 & 12.03 & 8.84            & \underline{3.06} & --- \\
    PTF~\cite{Wang2021PTF}           & 48.9 &  8.17 & \underline{8.68}  & 8.85            & 6.92            & --- \\
    RMR~\cite{Bhatnagar2019MGN}      & 43.2 &  \underline{7.80} & \textbf{8.60}  & 10.54           & 3.42            & --- \\
    Ours                              & \textbf{19.8} & \textbf{3.62} & 9.47  & \textbf{7.72}   & \textbf{2.62}   & \textbf{34.48} \\
    \bottomrule
  \end{tabular}
  }
  \caption{%
    \textbf{Quantitative comparison.}
    \emph{Body Fitting}: penetration rate, penetration depth, and scan proximity.
    \emph{Shape Est.}: bidirectional Chamfer distance of T-pose shape to minimum clothing shape in BuFF~\cite{Zhang2017BUFF}.
    \emph{Surf.\ Reg.}: bidirectional Chamfer distance between registration and scan.
    \emph{Tex.\ Reg.}: multiview rendering PSNR between textured
    registrations and original scans.
    Best in \textbf{bold}, second-best \underline{underlined}.
  }
  \label{tab:quantitative}
   \vspace{-2em}
\end{table}
\noindent\textbf{Body Fitting.}
We evaluate all methods that produce a naked body fit
(ETCH \cite{Li2025ETCH}, IPNet~\cite{Bhatnagar2020IPNet},
PTF~\cite{Wang2021PTF}, NICP~\cite{Marin2024NICP},
RMR~\cite{Bhatnagar2019MGN}, and ours) on the body fitting metrics
defined in \Cref{sec:eval_benchmark} across 4D-Dress (47 subjects),
BuFF (26) and CAPE (40).
\Cref{tab:quantitative} presents the results.
RMR and PTF achieve the lowest scan proximity
(8.60\,mm and 8.68\,mm), yet nearly
half of their body vertices penetrate the clothing surface
(43.3\% and 48.9\%, respectively).
In contrast, our method diminishes the penetration rate to
\textbf{19.8\%}, roughly half that of the next-best method
(ETCH, 35.0\%), with a penetration depth of only
\textbf{3.62\,mm} at maintain of competitive scan proximity
(9.47\,mm).
This demonstrates that our approach achieves a substantially better
trade-off between proximity and physical plausibility: the estimated
body sticks close to the clothing surface while remaining underneath
it.
Note that optimal proximity is not zero: a correctly estimated naked
body should maintain a physical offset from the outer clothing surface.
Crucially, our prevailing shape estimation on BuFF
\textbf{(7.72\,mm}, \cref{tab:quantitative}), where ground-truth
minimal-clothing body scans are available, confirms that our method
does not artificially shrink the body to avoid penetration, but
accurately recovers the true underlying body volume.

\noindent\textbf{Surface Registration.}
We measure clothed surface reconstruction via bidirectional Chamfer
distance (100k surface samples per mesh) between \smplx\smpld\
registration and input scan.
We compare IPNet~\cite{Bhatnagar2020IPNet}, PTF~\cite{Wang2021PTF},
NICP~\cite{Marin2024NICP}, RMR~\cite{Bhatnagar2019MGN}, and ours; ETCH is excluded as it produces only naked body
parameters.
As shown in \Cref{tab:quantitative} (\textit{Surf.\ Reg.}), our method achieves the lowest
Chamfer distance across all three datasets, with an overall mean of
\textbf{2.62\,mm}, a 14\% reduction over the
second-best method (NICP, 3.06\,mm).
Notably, our method also exhibits the lowest variance
($\sigma{=}0.39$\,mm), indicating consistently
accurate registrations irrespective of clothing type or body shape.
The improvement is most pronounced on 4D-Dress
(2.75\,mm ours vs.\ 3.53\,mm NICP),
which contains the most diverse clothing styles, suggesting that
our approach generalizes better to challenging garment geometries.
Template-based methods (PTF, IPNet) lag behind, likely as their
fixed clothing topology limits ability to conform to
diverse garment shapes.

\noindent\textbf{Texture Registration.}
\label{sec:exp_texture_mapping}
Beyond just geometry, our method produces complete textured registration
via remapping of scan appearance onto the \smplx\ UV layout.
To our knowledge, none of the publicly available baselines
(IPNet, PTF, NICP, RMR) support texture remapping, making a
direct comparison infeasible.
% Note that learning-based texture generators (\eg~\cite{Sanyal2024SCULPT,Kim2024PaintIt}) are orthogonal to our task: they \emph{synthesize} new appearance from prompts or partial observations, whereas \avaimg\ \emph{extracts} the actual scan texture into UV space, preserving the subject's true appearance rather than hallucinating it.
% We evaluate the fidelity of our textured registrations by rendering both the original scans and the registrations from multiple viewpoints using PyTorch3D~\cite{Ravi2020PyTorch3D}, and computing PSNR between paired renderings as our primary metric. We additionally report the Fr\'{e}chet Inception Distance (FID)~\cite{Heusel2017FID} as a distributional similarity measure.

As shown in \Cref{tab:quantitative} (\textit{Tex.\ Reg.}), our method
achieves a multiview rendering PSNR of \textbf{34.48\,dB} against
original scans, confirming the high visual fidelity of our textured
registrations.
In distributional terms, the FID between rendered registrations and
rendered scans is only \textbf{5.19}, indicating the two image
sets are statistically near-indistinguishable.
The low FID confirms our UV-based texture mapping preserves
fine appearance details (\eg fabric patterns, color gradients) with
high fidelity, despite the topological transformation from the scan's
native mesh to the \smplx\ UV parameterization.

\subsection{Image-based Avatar Representation}
\label{sec:vae_experiment}

Our successful realization of the Avatar-As-Image representation hinges upon a key condition: the attained UV maps as
produced by \avaimg\ being standardized 2D images, which pretrained generative
models can both represent and process imminently without modification.
We verify this by passing our UV texture and displacement maps through
the frozen VAE of FLUX~\cite{BlackForestLabs2024FLUX}---with no
fine-tuning---and then measuring the reconstruction fidelity in both UV image
space (PSNR, SSIM, LPIPS~\cite{Zhang2018LPIPS}) and 3D geometry space
(bidirectional Chamfer distance).

\begin{figure*}[t]
  \centering
  \includegraphics[width=\linewidth]{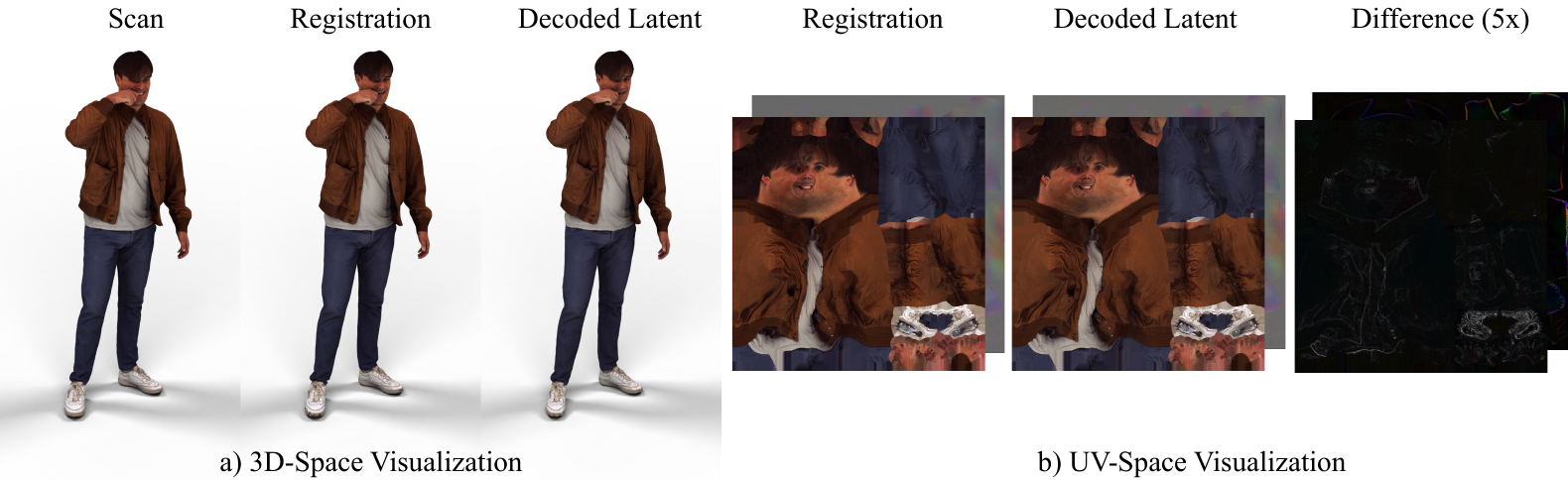}
  % [OLD — replaced by Claude 2026-03-04 (compressed caption)]
  % \textbf{(a)}~3D-space comparison: ... (10 lines → 4)
  \caption{%
    \textbf{VAE roundtrip validation.}
    (a)~Original scan, \avaimg\ registration, and mesh from VAE-decoded UV maps are visually near-indistinguishable.
    (b)~Original UV maps, VAE-reconstructed, and $5\times$ amplified difference.
  }
  \label{fig:vae_roundtrip}
\end{figure*}

\Cref{tab:vae} and \Cref{fig:vae_roundtrip} confirm that both UV maps
survive the roundtrip with minimal degradation.
Within 3D space, the VAE adds only \textbf{0.76\,mm} Chamfer error
(3.15\,mm $\to$ 3.91\,mm).
UV texture maps achieve a high \textbf{38.6\,dB} PSNR designating low texture atrophy, and displacement maps reconstruct
with \textbf{4.98\,mm} RMSE, well below the scale of feasibly represented clothing folds as per our supported topology. In spite of
the FLUX VAE being trained exclusively on natural images, we recognize only little geometric and visual error. 

% These results signify that \avaimg\ UV maps are a ready-to-use input format
% for image generative architectures, for which encoding-decoding have been shown to operate faithfully, and that they lie within natural-image distributions.
% through a frozen, off-the-shelf VAE with
% ?no retraining.
These results signify that \avaimg's UV maps lie within natural-image distributions and are a ready-to-use input format
for image generative architectures, for which encoding-decoding have been shown to operate faithfully.

\begin{table}[ht!]
  \centering
  \resizebox{\linewidth}{!}{%
  \setlength{\tabcolsep}{4pt}
  \begin{tabular}{l ccc c cccc}
    \toprule
    & \multicolumn{2}{c}{UV Texture} & UV Disp. & Spatial & \multicolumn{4}{c}{Spatial \& Rendering vs.\ Scan} \\
    \cmidrule(lr){2-3} \cmidrule(lr){4-4} \cmidrule(lr){5-5} \cmidrule(lr){6-9}
      & PSNR $\uparrow$ & SSIM $\uparrow$
      & RMSE (mm) $\downarrow$
      & V2V (mm) $\downarrow$
      & CD (mm) $\downarrow$ & PSNR $\uparrow$ & SSIM $\uparrow$ & LPIPS $\downarrow$ \\
    \midrule
    Before VAE  & ---   & ---   & ---  & ---  & 3.15 & 34.48 & 0.995 & 0.006 \\
    After VAE   & 38.6  & 0.964 & 4.98 & 4.29 & 3.91 & 30.04 & 0.988 & 0.009 \\
    \bottomrule
  \end{tabular}}
  \caption{%
    \textbf{VAE roundtrip fidelity}.
    UV texture and displacement maps as encoded and decoded through
    the frozen FLUX VAE~\cite{BlackForestLabs2024FLUX}
    vs.\ scan: bidirectional Chamfer distance and multiview rendering
    metrics (PSNR, SSIM, LPIPS), both measured against the original
    scan.
    The VAE roundtrip adds only 0.76\,mm Chamfer error.
  }
  \label{tab:vae}
 \vspace{-2em}
\end{table}

\subsection{Ablation Study}
\label{sec:ablation}

% [OLD — replaced by Claude 2026-03-01]
% \YX{Use signed distance or not for naked body register on 4DDress, BuFF, Cape}

We ablate three key design characteristics which constitute the \avaimg\ pipeline:
(1)~signed vs.\ unsigned body fitting (\cref{sec:body_fitting}),
(2)~coarse-to-fine refinement with fourth-power edge-coupling (\cref{sec:surface_reg}), and
(3)~efficient winding computation via decimation and winding band reduction (\cref{sec:body_fitting}).

\noindent\textbf{Signed vs.\ Unsigned Body Fitting.}
Our full method constrains body vertices to remain \emph{inside}
of clothing surface by incorporating sign to mesh-to-scan distance
via generalized winding numbers~\cite{Jacobson2013Winding}
(\cref{sec:body_fitting}), contrary to \eg RMR's~\cite{Bhatnagar2020RMR} employ of an unsigned metric (\cref{sec:relatedclothed}).
To evaluate the importance of this constraint, we compare against an
ablated variant which replaces the signed distance term with a
standard unsigned mesh-to-scan distance, identical to the data term
used by prior unsigned registration pipelines.

\begin{figure}[t]
  \centering
  \begin{minipage}[c]{0.42\linewidth}
    \centering
    \includegraphics[width=\linewidth]{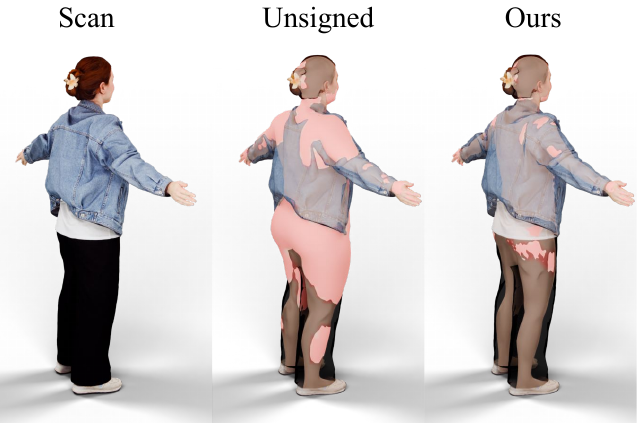}
  \end{minipage}
  \hfill
  \begin{minipage}[c]{0.55\linewidth}
    \centering
    \resizebox{\linewidth}{!}{%
    \setlength{\tabcolsep}{4pt}
    \begin{tabular}{l cccc}
      \toprule
      Distance
        & Pene.\ R.\ $\downarrow$
        & Pene.\ D.\ $\downarrow$
        & Prox.\ $\downarrow$
        & Shape$^\dagger$ $\downarrow$ \\
      \midrule
      Unsigned      & 40.7\% & 10.06\,mm & 9.74\,mm & 11.95\,mm \\
      Signed (ours) & \textbf{19.8\%} & \textbf{3.62}\,mm & \textbf{9.47}\,mm & \textbf{7.72}\,mm \\
      \bottomrule
      \multicolumn{5}{l}{\scriptsize $^\dagger$BuFF only (26 subj.; GT minimal body required).}
    \end{tabular}}
  \end{minipage}
  \caption{%
    \textbf{Ablation: signed vs.\ unsigned body fitting.}
    \emph{Left}: body mesh overlaid with the clothed scan; unsigned
    distance drives the body through the clothing surface
    (penetrating vertices in red), while our signed formulation
    keeps the body enclosed.
    \emph{Right}: quantitative comparison on 113 subjects
    (4D-Dress, BuFF, CAPE).
  }
  \label{fig:ablation_signed}
   \vspace{-2em}
\end{figure}

\Cref{fig:ablation_signed} presents the comparison.
Without signed winding numbers, the optimizer has no means to
distinguish whether the body approaches the scan surface from inside
or outside the clothing.
As a result, the body mesh frequently \emph{penetrates through} the
scan surface to minimize distance, achieving a deceptively low scan
proximity at the expense of physical plausibility.
With our signed distance formulation, penetration rate drops from
40.7\% to \textbf{19.8\%} and penetration depth from
10.06\,mm to \textbf{3.62\,mm}, while scan proximity remains
comparable (9.74\,mm vs.\ \textbf{9.47\,mm}).
The effect extends to shape estimation: on BuFF
where ground-truth minimal body shapes are available, shape-under-clothing
Chamfer distance drops from 11.95\,mm to \textbf{7.72\,mm}.
This confirms that the signed distance constraint is the primary
mechanism that prevents body--clothing collision and is essential for
physically plausible body estimation.

\noindent\textbf{Progressive Refinement.}
We ablate the two-pass coarse-to-fine displacement strategy
(\cref{sec:surface_reg}) by comparing three variants on the same
113-subject evaluation set:
(i)~\emph{low-res only}: displacement optimization of
original \smplx\smpld\ mesh without subdivision;
(ii)~\emph{two-pass, squared coupling}: applied Loop subdivision
followed by common squared
edge-coupling loss $(w_c\mathcal{L}_c)^2$; and
(iii)~our full method with fourth-power coupling
$(w_c\mathcal{L}_c)^4$ at high resolution.
% [OLD — replaced by Claude 2026-03-04 (filled in ablation numbers)]
% The low-res-only variant achieves a surface Chamfer distance of
% [X.XX]\,mm, ...
% \YX{TODO: fill in placeholder numbers from experiments}
The low-res only variant achieves a surface Chamfer distance of
3.15\,mm, as the limited vertex count cannot represent
high-frequency clothing detail.
Adding subdivision with squared coupling reduces the error to
2.75\,mm ($-$13\%), indicating that increased mesh resolution is
necessary but not sufficient.
Switching to fourth-power coupling (our full method) further
lowers the Chamfer to \textbf{2.62\,mm} ($-$5\%), as the wider penalty basin
allows the mesh to conform tighter to fine surface
structures instead of smoothing over them.
Body fitting and shape metrics are identical across all three
variants, as only the clothed surface registration
stage differs.

% [OLD — replaced by Claude 2026-03-04 (verbose decimation text + table moved to supp)]
% \paragraph{Decimation and Winding Band Acceleration Contribution.} \MK{TODO text}
\noindent\textbf{Decimation and Winding Band Reduction.}
We profiled the two acceleration components of our efficient winding
formulation (\cref{sec:body_fitting}) across decimation levels
with and without winding band reduction.
% (full table in supplementary material).
Most notably, decimating a select large scan of ${\sim}$$370$k faces down to 40k cuts winding computation
time from $\sim$$2{,}400\,$s to $\sim$$240$\,s (${\sim}$$10\times$).
The winding band reduction at cost of only $\sim$$70$\,s further accelerates body fitting from
${\sim}$$1{,}700$\,s to ${\sim}$$240$\,s (${\sim}$$7\times$), and facilitates strong storage saving via decreases of $\sim$$70$-$100$\,MB to $\sim$$5$\,MB~(up to $20\times$).
Combined, the physics-aware related stages drop from ${\sim}$$4{,}000$\,s
to ${\sim}$$540$\,s~(${\sim}7\times$) runtime with no quality degradation.

\section{Conclusion}
\label{sec:conclusion}

% [OLD — replaced by Claude 2026-03-04 (humanization pass + FID→PSNR fix)]
% We presented \avaimg, a multi-stage optimization pipeline that produces
% ... (FID\,=\,5.19) ...
% Most importantly, we showed that this registration quality is the missing
% enabler for the Avatar-As-Image representation ...
% ... \avaimg-quality UV maps is a natural next step.
We presented \avaimg, a multi-stage optimization pipeline that yields
high-fidelity \smplx\smpld\ registrations with UV texture mapping from
arbitrary clothed human scans at varying source qualities.
By enforcing strict body-inside-clothing containment via efficient signed winding
numbers per a three-level efficiency cascade ($\sim$10$\times$ runtime reduced, $\sim$95\% storage saved), 
our method effectively eliminates interpenetration artifacts common in existing
dataset ground truths and public registration tools.
Evaluations spanning six datasets confirm that \avaimg\ consistently outperforms all
state-of-the-art methods across body fitting, shape estimation, and surface registration---with 
textured registrations nearly indistinguishable from scans $(\text{PSNR}{=}34.48 \text{dB})$. 
Furthermore, \avaimg's resulting UV maps can be seamlessly encoded-decoded through a frozen image diffusion
VAE with only 0.76\,mm increased Chamfer error, 
% confirming that registration
% quality enables the Avatar-As-Image concept.
validating that the maps lie within natural-image distributions and that 
our registration pipeline successfully supports the Avatar-As-Image paradigm.
% enables the Avatar-As-Image paradigm.
% Code, data, and Singularity containers will be publicly released.

\vspace{\baselineskip}
\noindent\textbf{Limitations and Future Work.}
\avaimg's primary representational limitations stem from its base on \smplx\smpld, which bounds reconstructional ability on loose or decoupled garments 
and can result in proximity-based association errors in complex geometric regions. Self-contact zones are prone to texture bleeding, and unposing previously bent joints can introduce topological indentations. 
The full pipeline requires approximately $25$ minutes per scan, spanning 3D joint estimation, winding calculation, body and surface fitting, as well as texture mapping.
This defines it slower than feed-forward approaches, however, \avaimg\ rather targets one-time high-fidelity ground truth generation, which can serve as training data for faster alternatives.
Overcoming topological constraints and leveraging the Avatar-As-Image format for training of latent image diffusion models for 3D clothed humans remain promising directions for future work.

{\small
\subsubsection{\small Acknowledgments.} 
The authors express appreciation towards Yuliang Xiu and all others who gave any feedback towards improving this work.
This work is made possible by funding from the Carl Zeiss
Foundation. This work is also funded by the Deutsche Forschungsgemeinschaft (DFG, German
Research Foundation) - 409792180 (EmmyNoether Programme, project: Real Virtual Humans)
and the German Federal Ministry of Education and Research (BMBF): Tübingen AI Center, FKZ:
01IS18039A. The authors thank the International Max Planck Research School for Intelligent Systems
(IMPRS-IS) for supporting Y.Xue. G. Pons-Moll is a member of the Machine Learning Cluster of
Excellence, EXC number 2064/1 – Project number 390727645. 
}
{\small
\subsubsection{\small Disclosure of Interests.} 
The authors have no competing interests to declare that are
relevant to the content of this work.
}

% \subsubsection{\discintname}
% It is now necessary to declare any competing interests or to specifically
% state that the authors have no competing interests. Please place the
% statement with a bold run-in heading in small font size beneath the
% (optional) acknowledgments\footnote{If EquinOCS, our proceedings submission
% system, is used, then the disclaimer can be provided directly in the system.},
% for example: The authors have no competing interests to declare that are
% relevant to the content of this article. Or: Author A has received research
% grants from Company W. Author B has received a speaker honorarium from
% Company X and owns stock in Company Y. Author C is a member of committee Z.

% \clearpage  % TODO REVIEW/FINAL: This \clearpage needs to be removed from both review and camera-ready versions.

% ---- Supplementary material (combined build; see \ifincludesupplementary toggle above) ----

\bibliographystyle{splncs04}
\bibliography{main}

@String(PAMI  = {IEEE Trans. Pattern Anal. Mach. Intell.})

@String(CVPR  = {IEEE Conf. Comput. Vis. Pattern Recog.})

@String(ICCV  = {Int. Conf. Comput. Vis.})

@String(ECCV  = {Eur. Conf. Comput. Vis.})

@String(NeurIPS = {Adv. Neural Inform. Process. Syst.})

@String(BMVC  = {Brit. Mach. Vis. Conf.})

@String(TOG   = {ACM Trans. Graph.})

@String(PAMI  = {IEEE TPAMI})

@String(CVPR  = {CVPR})

@String(ICCV  = {ICCV})

@String(ECCV  = {ECCV})

@String(NeurIPS = {NeurIPS})

@String(BMVC  =	{BMVC})

@String(TOG   = {ACM TOG})

@article{Loper2015SMPL,
  author    = {Loper, Matthew and Mahmood, Naureen and Romero, Javier and Pons-Moll, Gerard and Black, Michael J.},
  title     = {{SMPL}: A Skinned Multi-Person Linear Model},
  journal   = TOG,
  volume    = {34},
  number    = {6},
  year      = {2015}
}

@inproceedings{Pavlakos2019SMPLX,
  author    = {Pavlakos, Georgios and Choutas, Vasileios and Ghorbani, Nima and Bolkart, Timo and Osman, Ahmed A. A. and Tzionas, Dimitrios and Black, Michael J.},
  title     = {Expressive Body Capture: {3D} Hands, Face, and Body from a Single Image},
  booktitle = CVPR,
  year      = {2019}
}

@inproceedings{Bogo2016SMPLify,
  author    = {Bogo, Federica and Kanazawa, Angjoo and Lassner, Christoph and Gehler, Peter and Romero, Javier and Black, Michael J.},
  title     = {Keep It {SMPL}: Automatic Estimation of {3D} Human Pose and Shape from a Single Image},
  booktitle = ECCV,
  year      = {2016}
}

@inproceedings{Bhatnagar2019MGN,
  author    = {Bhatnagar, Bharat Lal and Tiwari, Garvita and Theobalt, Christian and Pons-Moll, Gerard},
  title     = {Multi-Garment Net: Learning to Dress {3D} People from Images},
  booktitle = ICCV,
  year      = {2019}
}

@misc{Bhatnagar2020RMR,
  author       = {Bhatnagar, Bharat Lal},
  title        = {{RVH} Mesh Registration},
  year         = {2020},
  howpublished = {\url{https://github.com/bharat-b7/RVH_Mesh_Registration}},
  note         = {Accessed: 2026-03-02}
}

@inproceedings{Corona2022LVD,
  author    = {Corona, Enric and Pons-Moll, Gerard and Aleny{\`a}, Guillem and Moreno-Noguer, Francesc},
  title     = {Learned Vertex Descent: A New Direction for {3D} Human Model Fitting},
  booktitle = ECCV,
  year      = {2022}
}

@inproceedings{Feng2023ArtEq,
  author    = {Feng, Haiwen and Kulits, Peter and Liu, Shichen and Black, Michael J. and Abrevaya, Victoria Fern{\'a}ndez},
  title     = {Generalizing Neural Human Fitting to Unseen Poses with Articulated {SE}(3) Equivariance},
  booktitle = ICCV,
  year      = {2023}
}

@misc{Li2025ETCH,
  author    = {Li, Boqian and Feng, Haiwen and Cai, Zeyu and Black, Michael J. and Xiu, Yuliang},
  title     = {{ETCH}: Generalizing Body Fitting to Clothed Humans via Equivariant Tightness},
  year      = {2025},
  eprint    = {2503.10624},
  archiveprefix = {arXiv}
}

@inproceedings{Marin2024NICP,
  author    = {Marin, Riccardo and Corona, Enric and Pons-Moll, Gerard},
  title     = {{NICP}: Neural {ICP} for {3D} Human Registration at Scale},
  booktitle = ECCV,
  year      = {2024}
}

@inproceedings{Bhatnagar2020LoopReg,
  author    = {Bhatnagar, Bharat Lal and Sminchisescu, Cristian and Theobalt, Christian and Pons-Moll, Gerard},
  title     = {{LoopReg}: Self-Supervised Learning of Implicit Surface Correspondences, Pose and Shape for {3D} Human Mesh Registration},
  booktitle = NeurIPS,
  year      = {2020}
}

@inproceedings{Bhatnagar2020IPNet,
  author    = {Bhatnagar, Bharat Lal and Sminchisescu, Cristian and Theobalt, Christian and Pons-Moll, Gerard},
  title     = {Combining Implicit Function Learning and Parametric Models for {3D} Human Reconstruction},
  booktitle = ECCV,
  year      = {2020}
}

@inproceedings{Alldieck2019Tex2Shape,
  author    = {Alldieck, Thiemo and Pons-Moll, Gerard and Theobalt, Christian and Magnor, Marcus},
  title     = {{Tex2Shape}: Detailed Full Human Body Geometry From a Single Image},
  booktitle = ICCV,
  year      = {2019}
}

@article{Habermann2021DDC,
  author    = {Habermann, Marc and Liu, Lingjie and Xu, Weipeng and Zollh{\"o}fer, Michael and Pons-Moll, Gerard and Theobalt, Christian},
  title     = {Real-time Deep Dynamic Characters},
  journal   = TOG,
  volume    = {40},
  number    = {4},
  year      = {2021}
}

@inproceedings{Alldieck2018VideoAvatar,
  author    = {Alldieck, Thiemo and Magnor, Marcus and Xu, Weipeng and Theobalt, Christian and Pons-Moll, Gerard},
  title     = {Video Based Reconstruction of {3D} People Models},
  booktitle = CVPR,
  year      = {2018}
}

@inproceedings{Alldieck2018DetailedAvatars,
  author    = {Alldieck, Thiemo and Magnor, Marcus and Xu, Weipeng and Theobalt, Christian and Pons-Moll, Gerard},
  title     = {Detailed Human Avatars from Monocular Video},
  booktitle = {3DV},
  year      = {2018}
}

@inproceedings{Lazova2019360Tex,
  author    = {Lazova, Verica and Insafutdinov, Eldar and Pons-Moll, Gerard},
  title     = {360-Degree Textures of People in Clothing from a Single Image},
  booktitle = {3DV},
  year      = {2019}
}

@inproceedings{Lahner2018DeepWrinkles,
  author    = {L{\"a}hner, Zorah and Cremers, Daniel and Tung, Tony},
  title     = {{DeepWrinkles}: Accurate and Realistic Clothing Modeling},
  booktitle = ECCV,
  year      = {2018}
}

@inproceedings{Mir2020Pix2Surf,
  author    = {Mir, Aymen and Alldieck, Thiemo and Pons-Moll, Gerard},
  title     = {Learning to Transfer Texture from Clothing Images to {3D} Humans},
  booktitle = CVPR,
  year      = {2020}
}

@inproceedings{Chaudhuri2021SemiSupTex,
  author    = {Chaudhuri, Bindita and Sarafianos, Nikolaos and Shapiro, Linda and Tung, Tony},
  title     = {Semi-supervised Synthesis of High-Resolution Editable Textures for {3D} Humans},
  booktitle = CVPR,
  year      = {2021}
}

@article{Liu2021NeuralActor,
  author    = {Liu, Lingjie and Habermann, Marc and Rudnev, Viktor and Sarkar, Kripasindhu and Gu, Jiatao and Theobalt, Christian},
  title     = {Neural Actor: Neural Free-view Synthesis of Human Actors with Pose Control},
  journal   = TOG,
  volume    = {40},
  number    = {6},
  year      = {2021}
}

@inproceedings{Pang2024ASH,
  author    = {Pang, Haokai and Zhu, Heming and Kortylewski, Adam and Theobalt, Christian and Habermann, Marc},
  title     = {{ASH}: Animatable Gaussian Splats for Efficient and Photoreal Human Rendering},
  booktitle = CVPR,
  year      = {2024}
}

@article{Zhu2024TriHuman,
  author    = {Zhu, Heming and Zhan, Fangneng and Theobalt, Christian and Habermann, Marc},
  title     = {{TriHuman}: A Real-time and Controllable Tri-plane Representation for Detailed Human Geometry and Appearance Synthesis},
  journal   = TOG,
  volume    = {44},
  number    = {1},
  year      = {2024}
}

@inproceedings{Sun2025DUT,
  author    = {Sun, Guoxing and Dabral, Rishabh and Zhu, Heming and Fua, Pascal and Theobalt, Christian and Habermann, Marc},
  title     = {Real-time Free-view Human Rendering from Sparse-view {RGB} Videos using Double Unprojected Textures},
  booktitle = CVPR,
  year      = {2025}
}

@inproceedings{Casas2023SMPLitex,
  author    = {Casas, Dan and Comino-Trinidad, Marc},
  title     = {{SMPLitex}: A Generative Model and Dataset for {3D} Human Texture Estimation from Single Image},
  booktitle = BMVC,
  year      = {2023}
}

@inproceedings{Kim2024PaintIt,
  author    = {Kim, Youwang and Oh, Tae-Hyun and Pons-Moll, Gerard},
  title     = {Paint-It: Text-to-Texture Synthesis via Deep Convolutional Texture Map Optimization and Physically-Based Rendering},
  booktitle = CVPR,
  year      = {2024}
}

@inproceedings{Sanyal2024SCULPT,
  author    = {Sanyal, Soubhik and Ghosh, Partha and Yang, Jinlong and Black, Michael J. and Thies, Justus and Bolkart, Timo},
  title     = {{SCULPT}: Shape-Conditioned Unpaired Learning of Pose-Dependent Clothed and Textured Human Meshes},
  booktitle = CVPR,
  year      = {2024}
}

@inproceedings{Zheng2019DeepHuman,
  author    = {Zheng, Zerong and Yu, Tao and Wei, Yixuan and Dai, Qionghai and Liu, Yebin},
  title     = {{DeepHuman}: {3D} Human Reconstruction from a Single Image},
  booktitle = ICCV,
  year      = {2019}
}

@inproceedings{Yu2021Function4D,
  author    = {Yu, Tao and Zheng, Zerong and Guo, Kaiwen and Liu, Pengpeng and Dai, Qionghai and Liu, Yebin},
  title     = {{Function4D}: Real-time Human Volumetric Capture from Very Sparse Consumer {RGBD} Sensors},
  booktitle = CVPR,
  year      = {2021}
}

@inproceedings{Han20232K2K,
  author    = {Han, Sang-Hun and Park, Min-Gyu and Yoon, Ju Hong and Kang, Ju-Mi and Park, Young-Jae and Jeon, Hae-Gon},
  title     = {High-Fidelity {3D} Human Digitization from Single {2K} Resolution Images},
  booktitle = CVPR,
  year      = {2023}
}

@inproceedings{Ho2023CustomHumans,
  author    = {Ho, Hsuan-I and Xue, Lixin and Song, Jie and Hilliges, Otmar},
  title     = {Learning Locally Editable Virtual Humans},
  booktitle = CVPR,
  year      = {2023}
}

@article{Cao2021OpenPose,
  author    = {Cao, Zhe and Hidalgo Martinez, Gines and Simon, Tomas and Wei, Shih-En and Sheikh, Yaser},
  title     = {{OpenPose}: Realtime Multi-Person {2D} Pose Estimation Using Part Affinity Fields},
  journal   = PAMI,
  volume    = {43},
  number    = {1},
  pages     = {172--186},
  year      = {2021}
}

@inproceedings{Ravi2020PyTorch3D,
  author    = {Ravi, Nikhila and Reizenstein, Jeremy and Novotny, David and Gordon, Taylor and Lo, Wan-Yen and Johnson, Justin and Gkioxari, Georgia},
  title     = {Accelerating {3D} Deep Learning with {PyTorch3D}},
  booktitle = {SIGGRAPH Asia Courses},
  year      = {2020}
}

@article{Jacobson2013Winding,
  author    = {Jacobson, Alec and Kavan, Ladislav and Sorkine-Hornung, Olga},
  title     = {Robust Inside-Outside Segmentation Using Generalized Winding Numbers},
  journal   = TOG,
  volume    = {32},
  number    = {4},
  year      = {2013}
}

@inproceedings{Tang2025HGD,
  author    = {Tang, Xiangjun and Zhang, Biao and Wonka, Peter},
  title     = {Generative Human Geometry Distribution},
  booktitle = CVPR,
  year      = {2025}
}

@misc{BlackForestLabs2024FLUX,
  author       = {{Black Forest Labs}},
  title        = {{FLUX}.1},
  year         = {2024},
  howpublished = {\url{https://github.com/black-forest-labs/flux}}
}

@inproceedings{Wang2021PTF,
  author    = {Wang, Shaofei and Geiger, Andreas and Tang, Siyu},
  title     = {Locally Aware Piecewise Transformation Fields for {3D} Human Mesh Registration},
  booktitle = CVPR,
  pages     = {7639--7648},
  year      = {2021}
}

@inproceedings{Wang20244DDress,
  author    = {Wang, Wenbo and Ho, Hsuan-I and Guo, Chen and Rong, Boxiang and Grigorev, Artur and Song, Jie and Zarate, Juan Jose and Hilliges, Otmar},
  title     = {{4D-DRESS}: A {4D} Dataset of Real-World Human Clothing With Semantic Annotations},
  booktitle = CVPR,
  pages     = {550--560},
  year      = {2024}
}

@inproceedings{Zhang2017BUFF,
  author    = {Zhang, Chao and Pujades, Sergi and Black, Michael J. and Pons-Moll, Gerard},
  title     = {Detailed, Accurate, Human Shape Estimation from Clothed {3D} Scan Sequences},
  booktitle = CVPR,
  pages     = {4191--4200},
  year      = {2017}
}

@article{Pons-Moll2015Dyna,
  author    = {Pons-Moll, Gerard and Romero, Javier and Mahmood, Naureen and Black, Michael J.},
  title     = {Dyna: A Model of Dynamic Human Shape in Motion},
  journal   = TOG,
  volume    = {34},
  number    = {4},
  pages     = {120:1--120:14},
  year      = {2015}
}

@inproceedings{Pons-Moll2017ClothCap,
  author    = {Pons-Moll, Gerard and Pujades, Sergi and Hu, Sonny and Black, Michael J.},
  title     = {{ClothCap}: Seamless {4D} Clothing Capture and Retargeting},
  booktitle = TOG,
  volume    = {36},
  number    = {4},
  pages     = {73:1--73:15},
  year      = {2017}
}

@inproceedings{Ma2020CAPE,
  author    = {Ma, Qianli and Yang, Jinlong and Ranjan, Anurag and Pujades, Sergi and Pons-Moll, Gerard and Tang, Siyu and Black, Michael J.},
  title     = {Learning to Dress {3D} People in Generative Clothing},
  booktitle = CVPR,
  pages     = {6468--6477},
  year      = {2020}
}

@inproceedings{Heusel2017FID,
  author    = {Heusel, Martin and Ramsauer, Hubert and Unterthiner, Thomas and Nessler, Bernhard and Hochreiter, Sepp},
  title     = {{GANs} Trained by a Two Time-Scale Update Rule Converge to a Local Nash Equilibrium},
  booktitle = NeurIPS,
  pages     = {6626--6637},
  year      = {2017}
}

@inproceedings{Yan2025Omages,
  author    = {Yan, Xingguang and Lee, Han-Hung and Wan, Ziyu and Chang, Angel X.},
  title     = {An Object is Worth 64x64 Pixels: Generating {3D} Object via Image Diffusion},
  booktitle = {3DV},
  year      = {2025}
}

@inproceedings{Zhang2018LPIPS,
  author    = {Zhang, Richard and Isola, Phillip and Efros, Alexei A. and Shechtman, Eli and Wang, Oliver},
  title     = {The Unreasonable Effectiveness of Deep Features as a Perceptual Metric},
  booktitle = CVPR,
  year      = {2018}
}

@article{loop1987smooth,
  title={Smooth subdivision surfaces based on triangles},
  author={Loop, Charles},
  year={1987}
}

@inproceedings{Xue2026GeoRelight,
  author    = {Xue, Yuxuan and Liang, Ruofan and Zakharov, Egor and Bagautdinov, Timur and Cao, Chen and Nam, Giljoo and Saito, Shunsuke and Pons-Moll, Gerard and Romero, Javier},
  title     = {{GeoRelight: Learning Joint Geometrical Relighting and Reconstruction with Flexible Multi-Modal Diffusion Transformers}},
  booktitle = CVPR,
  year      = {2026},
}

@inproceedings{Xue2025InfiniHuman,
  author    = {Xue, Yuxuan and Xie, Xianghui and Kostyrko, Margaret and Pons-Moll, Gerard},
  title     = {{InfiniHuman: Infinite 3D Human Creation with Precise Control}},
  booktitle = {SIGGRAPH Asia 2025 Conference Papers},
  year      = {2025},
}

@article{Xue2025Gen3Diffusion,
  author    = {Xue, Yuxuan and Xie, Xianghui and Marin, Riccardo and Pons-Moll, Gerard},
  title     = {{Gen-3Diffusion: Realistic Image-to-3D Generation via 2D \& 3D Diffusion Synergy}},
  journal   = PAMI,
  year      = {2025},
  doi       = {10.1109/TPAMI.2025.3577067},
}

@inproceedings{Xue2024Human3Diffusion,
  author    = {Xue, Yuxuan and Xie, Xianghui and Marin, Riccardo and Pons-Moll, Gerard},
  title     = {{Human-3Diffusion: Realistic Avatar Creation via Explicit 3D Consistent Diffusion Models}},
  booktitle = NeurIPS,
  year      = {2024},
}

@inproceedings{Xue2023NSF,
  author    = {Xue, Yuxuan and Bhatnagar, Bharat Lal and Marin, Riccardo and Sarafianos, Nikolaos and Xu, Yuanlu and Pons-Moll, Gerard and Tung, Tony},
  title     = {{NSF: Neural Surface Fields for Human Modelling from Monocular Depth}},
  booktitle = ICCV,
  year      = {2023},
}

@inproceedings{wang2026dirtymocap,
    title={DirtyMoCap: Robust Motion Capture from Unconstrained Markers},
    author={Long Wang and Shuting Zhao and Shen Yan and Siyuan Yu and Xiaoben Li and Zeyu Cai and Yumeng Hou and Yuliang Xiu},
    booktitle={SIGGRAPH Asia 2026 Conference Papers},
    year={2026},
}

@inproceedings{cai2026omnifit,
      title={OmniFit: Multi-modal 3D Body Fitting via Scale-agnostic Dense Landmark Prediction},
      author={Zeyu Cai and Yuliang Xiu and Renke Wang and Zhijing Shao and Xiaoben Li and Siyuan Yu and Chao Xu and Yang Liu and Baigui Sun and Jian Yang and Zhenyu Zhang},
      booktitle={European Conference on Computer Vision (ECCV)},
      month={September},
      year={2026},
}

@inproceedings{li2026etchx,
    title={ETCH-X: Robustify Body Fitting to Any Clothed Human Scans with Composable Synthetic Data},
    author={Xiaoben Li and Jingyi Wu and Zeyu Cai and Siyuan Yu and Boqian Li and Yuliang Xiu},
    booktitle={European Conference on Computer Vision (ECCV)},
    month={September},
    year={2026},
}
\end{document}